\documentclass[preprint,12pt]{elsarticle}

\usepackage{amssymb}
\usepackage{amsmath}
\usepackage{algorithm}
\usepackage{algorithmic}
\usepackage{colortbl}
\usepackage{makecell}

\usepackage{subcaption}

\usepackage{booktabs}
\usepackage{tabularx}

\journal{Neurocomputing}

\begin{document}

\begin{frontmatter}


\title{KnowEEG: Explainable Knowledge Driven EEG Classification}

%% Authors with explicit affiliation labels
\author[inst1]{Amarpal Sahota\corref{cor1}}
\ead{amarpal.sahota@bristol.ac.uk}

\author[inst2]{Navid Mohammadi Foumani}
\ead{Navid.Foumani@monash.edu}

\author[inst1]{Raul Santos-Rodriguez}
\ead{enrsr@bristol.ac.uk}

\author[inst1]{Zahraa S.~Abdallah}
\ead{zahraa.abdallah@bristol.ac.uk}

\cortext[cor1]{Corresponding author.}

%% Affiliations
\affiliation[inst1]{%
    organization={University of Bristol},
    city={Bristol},
    country={United Kingdom}
}

\affiliation[inst2]{%
    organization={Monash University},
    city={Melbourne},
    state={Victoria},
    country={Australia}
}

%% Abstract
\begin{abstract}
Electroencephalography (EEG) classification has shown significant promise in applications ranging from neurodegenerative disease detection to brain-computer interfaces (BCIs). Whilst deep learning methods have increasingly dominated the research landscape in recent years; these models remain largely opaque ‘black boxes’ with limited interpretability. In this paper, we revisit the utility of feature-based classification via a systematic pipeline designed to integrate per-electrode signal statistics with cross-brain connectivity metrics. The proposed "KnowEEG" pipeline extracts 783 per-electrode statistical features alongside between-electrode connectivity statistics, utilizing a balanced sampling strategy between high-dimensional per electrode features and connectivity.
This feature-based pipeline achieves performance comparable to or exceeding state-of-the-art deep learning models across five diverse classification tasks: emotion detection, mental workload classification, eyes open/closed detection, abnormal EEG classification, and event detection.  This performance is achieved without the need for Graphics Processing Unit (GPU) resources. Furthermore, the pipeline provides inherent explainability through feature importance rankings. We demonstrate via the eyes open/closed task that this explainability identifies discriminative biomarkers in the occipital region that are consistent with established neuroscience literature. These results suggest that despite significant advances in deep learning, structured feature engineering remains a high performance and interpretable method for EEG classification.
\end{abstract}

% %% Research highlights 
% \begin{highlights}
% \item Propose KnowEEG, an explainable feature-based pipeline for EEG classification.
% \item Combine 10k+ per-electrode time-series features with connectivity metrics via a Fusion Forest.
% \item Match or exceed state-of-the-art deep learning models on five diverse EEG datasets.
% \item Provide inherent explainability that recovers known neurophysiological patterns.
% \end{highlights}

% %% Keywords
% \begin{keyword}
% EEG classification \sep Explainable AI \sep Time series features \sep Connectivity measures \sep Random forest 
% \end{keyword}

\end{frontmatter}

\section{Introduction}
\label{intro}

% Electroencephalogram (EEG) is a widely used method of recording cortical neuronal activity with a high temporal resolution \cite{casson2018electroencephalogram}.  Data collection is relatively inexpensive and also non-invasive. Thus, EEG has great potential to drive advancements in mental health, enable sophisticated Brain-Computer Interfaces (BCIs), and enhance disease detection, paving the way for future innovations in healthcare. Extensive research has explored EEG classification with applications ranging from the classification of neurodegenerative diseases such as Alzheimer's \cite{sahota2024interpretable} to motor task recognition for Brain-Computer Interfaces \cite{abiri2019comprehensive} and emotion detection \cite{katsigiannis2017dreamer}. 

Electroencephalography (EEG) provides a non-invasive and cost-effective means of recording cortical neuronal activity with high temporal resolution \cite{casson2018electroencephalogram}. Its accessibility drives its potential for healthcare innovation, particularly in advancing mental health diagnostics, enabling sophisticated brain-computer interfaces (BCIs) and enhancing disease detection. Existing research has explored EEG classification across diverse tasks, ranging from the identification of neurodegenerative disease like Parkinson’s \cite{sahota2024interpretable} to motor task recognition and visual target detection for BCIs \cite{abiri2019comprehensive} \cite{shi2024low_target_detect} and emotion detection \cite{katsigiannis2017dreamer}.

% As a non-invasive and inexpensive modality, electroencephalography (EEG) allows for the recording of cortical neuronal activity with a high temporal resolution \cite{casson2018electroencephalogram}. These characteristics underpin its potential to impact healthcare through advancements in mental health, sophisticated brain-computer interfaces (BCIs), and enhanced disease detection \cite{casson2018electroencephalogram}. Research in EEG classification has already demonstrated success across varied applications ranging from the detection of neurodegenerative diseases such as Parkinson's \cite{sahota2024interpretable} to motor task recognition for BCIs \cite{abiri2019comprehensive}, and emotional state classification \cite{katsigiannis2017dreamer}.

%(Band power, entropy, spectral features) followed by a machine learning model such as a support vector machine, random forest or artificial neural network to the direct application of a deep learning model such as a convolutional neural network. See section \ref{related work} for a detailed review. 

EEG classification is a challenging task due to the low signal-to-noise ratio in EEG signals and the presence of significant inter-subject variability. Traditional machine learning methods for EEG classification have relied on domain knowledge, defined here as the application of neuroscientific expertise to select features known to be informative of brain activity, such as signal power across specific frequency bands. These hand-crafted features are typically used as inputs to models such as support vector machines, random forests, or neural networks \cite{chaturvedi2017quantitative} \cite{betrouni2019electroencephalography} \cite{sahota2024interpretable} . We categorize such pipelines as 'knowledge-driven' because they utilize pre-defined mathematical descriptors rather than automated feature discovery. While these methods benefit from inherent interpretability, a key limitation is the bias induced by the selected input features. In settings where the pre-calculated features are not informative for classification, poor performance follows. For this reason, research in the last decade has focused increasingly on deep learning methodologies.

% Domain knowledge, specifically refers to expert knowledge about the task and the features likely to be informative for that task. In the context of EEG for example, it is well studied that signal power across frequency bands is informative of brain activity. 

Deep learning methods in this context range from convolutional neural networks (CNN) applied to the raw EEG data to audio-inspired networks that apply a CNN to EEG spectrograms (time-frequency image of the EEG) \cite{oh2020deep}. Recent advances in deep learning architectures have also led to transformer-inspired networks such as EEG Conformer \cite{eegconformer} and EEG2Rep \cite{mohammadi2024eeg2rep} (self-supervised representation-based model) for classification. Although the performance across EEG classification tasks has improved, deep learning models are often opaque and their inner workings are hard to interpret. This leads researchers to post-hoc explainability methods in an attempt to understand models \cite{posthoc_2022}. Explainability is a particularly critical issue for EEG as it can enhance insights into neurophysiological phenomena, aid clinical decision-making and improve trust in the model.

Therefore, while data and compute hungry deep learning methods also have issues around explainability, existing feature-based methods have a perceived inability to achieve high classification performance across diverse EEG classification tasks. In this paper, we investigate whether a feature-based pipeline with a diverse set of global EEG features can achieve classification performance comparable to state of the art deep learning models across five different task domains.

\begin{figure*}[ht!]
        \centering
        \includegraphics[width=1.0\textwidth]{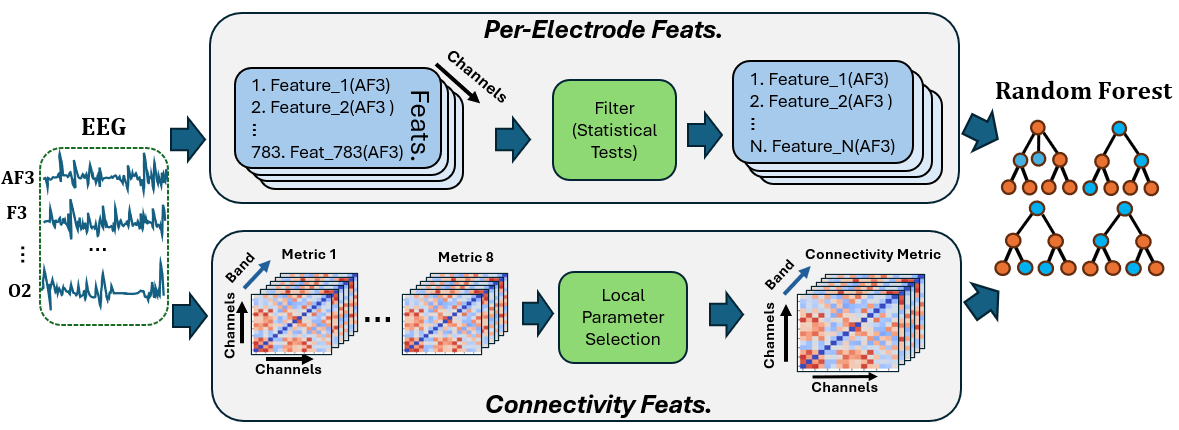}
        \caption{Feature Based Pipeline KnowEEG : There are two parallel threads to the pipeline. In thread one, 783 features from generalised and EEG-specific time-series literature are calculated per electrode and concatenated. Feature relevance is evaluated via univariate non-parametric hypothesis tests (Mann-Whitney U for binary targets and Kruskal-Wallis tests for multi-class targets) using only the training split of the data. The Benjamini–Yekutieli correction to control the false discovery rate is applied ($\alpha = 0.05$) with uninformative features filtered out. In thread two, between-electrode connectivity features are calculated for connectivity metrics. The best-performing metric is selected using classifier performance on the validation data. The features from thread one and thread two are then combined using a balanced sampling strategy (see Algorithm~\ref{alg:fusionforest}).
}
        \label{pipeline_fig}
\end{figure*}

For the feature based pipeline we leverage existing time series and EEG literature to construct a large feature space of 783 per electrode statistics that are filtered and then combined with between electrode connectivity statistics \cite{christ2018time} \cite{sahota2024interpretable}. These features are input to a modified Random Forest model that balances per electrode statistics with between electrode connectivity features in growing the trees of the forest (see pipeline in Figure \ref{pipeline_fig} and balanced sampling, Algorithm \ref{alg:fusionforest}). To facilitate consistent benchmarking across diverse datasets, we denote this systematic configuration of features and balanced sampling as the KnowEEG pipeline.

In this paper, we present the feature based pipeline KnowEEG in Section \ref{method}. In Section \ref{results}, we present the results of this pipeline versus state-of-the-art competitors across five different EEG classification task domains (Emotion detection, mental workload classification, eyes open/closed detection, Abnormal EEG classification, Event Detection), showing the feature based pipeline to be competitive with state of the art performance across tasks. Finally, in Section \ref{analysis}, we demonstrate the explainability benefits of utilising a feature based pipeline via analysis of feature importance for the eyes open vs. eyes closed classification task.

\section{Related Work}
In this section, we present related work which includes feature-based methods for general time series classification (across domains), feature based methods for EEG classification, and deep learning for EEG classification.

\subsection{Feature Engineering for General Time Series}

Time series data covers vastly different domains from energy consumption to EEG and weather prediction. Although different, all domains share the fundamental property of sequentially ordered observations that change over time.
Basic statistical descriptors such as mean, variance, skewness and kurtosis are often used as a starting point for classification. Features can then be extended to temporal patterns which include autocorrelation at different lags, peak detection and peak counting among others. In \cite{lubba2019catch22}, the authors propose a set of 22 features for generalized time series classification after extensive testing on 93 time series datasets. The `catch22' features include linear and non-linear autocorrelation, successive differences, value distributions and outliers and fluctuation scaling properties \cite{lubba2019catch22}. Features for generalized time series classification can also include shape-based features such as shapelets or motifs that capture local patterns. More recent work in this direction incorporates the semantic relationships between sub-series, for example by embedding discretised sub-series and comparing them with a deep graph kernel \cite{yu_et_al_2024}.

Time series statistical descriptors can be very effective for specific EEG classification tasks. For example researchers in \cite{4_recommended_rehman2023eeg} achieved a classification accuracy of over 95\% using 23 statistical features for depression classification on the MODMA dataset \cite{modma_dataset}. These features ranged from kurtosis and standard deviation to signal entropy, energy and more derived peak to peak signal measures. In \cite{3_recommended_khan2024machine} researchers demonstrated that a more concise set of only 12 features could achieve a classification accuracy of 96\% on the depression classification task.

While these general purpose statistics capture fundamental signal dynamics, they can be further refined through specialized EEG measures. These include connectivity and frequency-based metrics which we discus in detail in the subsequent section \ref{feat_eeg}.

\subsection{Feature Engineering for EEG} \label{feat_eeg}

Research has shown frequency-based features to be particularly informative of brain activity \cite{bandpowerpaper} and therefore effective for EEG classification. The power spectral density (strength of the EEG signal across frequencies) is calculated using the Fourier transform. From the power spectral density band powers and peak frequency values can be calculated. Band powers refer to specifically defined frequency bands. For our paper we use the band power definitions as delta (0.5--4Hz), theta (4--8Hz), alpha (8--12Hz), sigma (12--16Hz), beta (16--30Hz) and gamma (30--40Hz). Band power values and peak / median frequency features can be input to machine learning models such as support vector machines or random forests for classification as in \cite{chaturvedi2017quantitative} and \cite{betrouni2019electroencephalography}. 

The afore-discussed features, though effective, reduce temporal resolution to the window over which the feature was calculated and discard phase information within the signal. Connectivity measures are metrics for calculating connectivity between regions of the brain and often utilize phase information. Phase Lag Index (PLI) for example \cite{kuang2022phase}, measures the asymmetry of the distribution of phase differences across time between two signals with its value ranging between 0 and 1. A PLI of zero means no coupling (or instantaneous coupling) and a value of 1 means true lagged interaction. Connectivity measures can be calculated for the entire EEG signal between electrodes or for sub-components of the EEG signal within the frequency domain (commonly band powers). For example in \cite{kuang2022phase}, researchers find the phase lag index in the alpha band to be correlated with cognitive assessment score in the task of Mild Cognitive Impairment classification. 

Brain connectivity in general comprises the following subcomponents; structural connectivity (anatomical connectivity between brain regions), functional connectivity (statistical dependencies or correlations in activity) and effective connectivity (causal or directional influences between regions) \cite{chiarion2023connectivity}. Another commonly used metric is coherence which measures functional connectivity via correlation in the frequency domain. In \cite{waninger2020neurophysiological} researchers use coherence combined with other signal statistics to achieve over 90\% accuracy in the task of Parkinson's classification. 

Thus, in summary, feature-based approaches inspired by domain knowledge have been successful for EEG classification often focusing on frequency-based properties of the signal and measures of connectivity across the brain. 

% While many feature-engineering approaches exist, none provide a unified, high-performance, explainable pipeline applicable across multiple EEG classification domains.

\subsection{Deep Learning for EEG} \label{deeplearning}

Over the last decade, deep learning has experienced unprecedented growth and success, with complex neural architectures becoming the state of the art in numerous domains from natural language processing and autonomous systems to facial expression recognition and medical image reconstruction / super-resolution \cite{noor2025survey_deep_learning} \cite{medical_image_super_resolution} . This has led to much research experimenting with existing deep learning architectures applied to EEG data and the development of EEG-specific deep neural architectures \cite{mohammadi2024eeg2rep}, spanning the full EEG processing pipeline from artifact removal \cite{zhang2026stand} through to classification tasks such as emotion recognition \cite{ye2024adaptive} and brain–computer interfaces for rehabilitation \cite{chen2025novel}.

% Removed this 'low light, facial reconstruction' reference \cite{recommended_facial_recognition_xie2025joint} 
% Over the last decade, deep learning has experienced unprecedented growth and success across domains. This has led to much research experimenting with existing deep learning architectures applied to EEG data and the development of EEG-specific deep neural architectures\cite{mohammadi2024eeg2rep}. 

Firstly, researchers have effectively applied simple convolutional neural networks (CNNs) directly to processed EEG data for classification.  In \cite{oh2020deep} a CNN comprising of a series of 1D-convolutions and max pooling layers achieved just under 90\% accuracy on Parkinson's classification. EEG researchers have since taken inspiration from the audio domain by applying CNNs to time-frequency images (spectrograms) of the raw EEG signal \cite{li2022deep} \cite{mandhouj2021automated} \cite{1_recommended_khan2025transfer} \cite{2_recommended_javed2024deep}. Alternative image-based approaches are explored in papers such \cite{chen2026brain} where the authors instead encode the spatial distribution of band power across the scalp, projecting electrode positions into 2D and interpolating to form topographic power maps as CNN input.

Deep learning architectures have also been developed to deal directly with sequential data from RNN variants such as LSTMs and GRU's to more recently Transformers. Transformers excel at capturing long-range dependencies and as such have been successfully used for EEG classification. In \cite{eegconformer} the authors present a novel architecture that uses a convolutional module to generate embeddings for input EEG data which is then passed to a Transformer Encoder module for classification. This EEG Conformer achieves state of the art performance on many brain computer interface based tasks \cite{eegconformer}. 

One issue with deep neural networks is susceptibility to over-fitting \cite{mohammadi2024eeg2rep}. With EEG data this is particularly problematic as signals are noisy and inter-subject variability can be present \cite{schirrmeister2017deep}. Self-supervised learning (SSL) aims to solve this problem by learning a representation of the EEG and generating a self-supervisory signal from the data \cite{weng2024self}. These models can learn from labelled or unlabeled data and the learnt representations used for classification. State-of-the-art SSL deep learning models include BENDR \cite{kostas2021bendr}, BIOT \cite{yang2024biot} and most recently EEG2Rep \cite{mohammadi2024eeg2rep}.

Another limitation of deep neural networks is the sheer volume of training data required to achieve high levels of performance. Yue et al. \cite{yue2025augmenting} address this by introducing a neurophysiologically grounded data augmentation method that isolates background noise from one class and combines it with the task-related template of another. This technique yielded significant performance gains across multiple deep learning architectures, most notably enabling a Transformer-based model to achieve state-of-the-art results on a steady-state visually evoked potential (SSVEP) classification task.

Thus, in summary, many deep learning models have been applied to and developed specifically for EEG data. Deep learning has been successful in the classification of EEG. However, explainability remains an issue with deep learning models being opaque.

\section{Methodology of Feature Based Pipeline KnowEEG} \label{method}

\subsection{Problem Statement}
We address the problem of EEG classification. Each EEG sample $X_i$ from a dataset $\{X_1, X_2, \dots, X_N\}$ maps to a corresponding label $y_i$ from the set $\{y_1, y_2, \dots, y_N\}$. The label is a scalar value corresponding to the class of each respective sample.  Each EEG sample $X_i$ is multi-dimensional consisting of K electrode channels each with a sequence length of L. 

Our goal is to learn an explainable high-performance classifier that can map samples $X_i$ to labels  $y_i$. This is primarily measured with performance. Accuracy and Area Under the Receiver Operating Characteristic Curve (AUROC) are used as performance metrics for binary classification. Balanced Accuracy and Weighted F1 Score are used for multi-class classification. Explainability is established via generating understandable features and using a tree-based model that provides direct access to feature importances. Explainability is explored in \ref{analysis}.

\subsection{KnowEEG Model Architecture}

% \paragraph{Thread One : Per electrode features} \
\subsubsection*{\textbf{Thread One : per electrode features.}}
Per electrode we calculate 783 features from both the generalized time series domain and the EEG-specific domain. This includes basic features like mean, standard deviation and kurtosis to more complex time series features like autocorrelation at different lags and EEG specific features related to frequency components present in the signal. In order to do this calculation the time series Python package TSFresh \cite{christ2018time} is used. Exact settings for feature calculation are presented in \ref{experimental_procedure} and further details for implementation provided in \ref{appendix_knowEEG}. These per electrode features are concatenated after being calculated. Therefore, for our 14 electrode datasets there are 783 features x 14 electrodes which equals 10,962 time series features per sample (12,528 for 16 electrode datasets). Our aim is to create a broad set of time series descriptors for each EEG sample. We hypothesize that this broad set of features contains the subset of informative features for the specific EEG classification task. To avoid model over-fitting and reduce unnecessary computation, features are filtered using univariate non-parametric hypothesis tests to assess their relevance to the classification target. For binary classification we use the Mann-Whitney U test while for multi-class classification Kruskal–Wallis tests are applied. Resulting p-values are adjusted to control the false discovery rate using the Benjamini-ekutieli correction, with a significance threshold of 0.05 \cite{benjamini2001control}. This filtering and calculation of p values is done using only the training split of the data. Thus, the per electrode thread of the pipeline results in a final set of filtered per electrode features as shown in Figure \ref{pipeline_fig}.

% We select Spearman and Pearson correlation, which measure linear and non-linear correlation between electrodes.

% We select Pearson and Spearman correlation, which measure linear and monotonic rank-based association between electrodes respectively.

% \paragraph{Thread Two : Between electrode connectivity features:}
\subsubsection*{\textbf{Thread Two: between electrode connectivity features.}} We calculate a separate set of between electrode connectivity features. We select Pearson and Spearman correlation, which measure linear and monotonic rank-based association between electrodes respectively. For a 14 electrode dataset this results in 182 correlation features (91 Spearman and 91 Pearson). We also select a series of well researched EEG specific connectivity measures. For our pipeline we select Coherence (Coh) , Imaginary Coherence (ImCoh), Pairwise Phase Consistency (PPC), Phase Locking Value (PLV), Phase Lag Index (PLI) , Directed Phase Lag Index (DPLI) and Weighted Phase Lag Index (WPLI) \cite{MNE_2013}. Users of the pipeline can select any number of connectivity metrics that they expect could be informative for classification. These connectivity metrics can be calculated for the entire signal or on sub components of the signal. We propose calculating these metrics for each sub band of the EEG signal with regards to the bands defined in \ref{feat_eeg}. Therefore per connectivity metric, for the 14 electrode datasets there are 91 (features per sub band) x 6 sub bands (delta, theta, alpha, sigma, beta, gamma) which equates to 546 total features per metric (720 features for 16 electrode datasets). These connectivity metrics require segmentation of the raw signal into epochs prior to calculation in order to measure the temporal synchronisation of neural oscillations. We follow standard protocol for segmentation and detail this in \ref{appendix_connectivity}. In addition to these conventional connectivity metrics we also define Functional Power Connectivity (FPC) which consists of per electrode relative band powers and electrode-electrode features capturing spatial differences in oscillatory power across brain regions. Unlike phase-based metrics, FPC can be computed reliably even from short signals where windowing introduces noise. FPC is an amplitude-based measure that interprets functional connectivity through the spatial distribution and relative differences in oscillatory power across brain regions, calculated specifically as the pairwise log-ratio of relative bandpower within defined frequency bands (see \ref{FPC} for full FPC definition). Therefore, in total we have 9 candidate connectivity metrics.

We consider selection of a single connectivity metric in this thread of the pipeline as a hyperparameter. We propose selecting a single connectivity metric as opposed to selecting some features from all metrics for ease of interpretability. This hyperparameter is selected using 'local parameter selection' in order to reduce  the computational requirement of the  KnowEEG pipeline. Local parameter selection means that only the connectivity data is being used for parameter selection and not the per electrode statistics. A default 100 tree Random Forest model is used to select the highest performing connectivity metric on the validation set, fitting to each connectivity metric one at a time and testing performance on the validation set as shown in Figure \ref{pipeline_fig} . 

% \paragraph{Feature Fusion: Per-electrode and between electrode features}
\subsubsection*{\textbf{Feature Fusion: per-electrode and between electrode features.}}
We now have per electrode features from thread one and between electrode features (connectivity measure) from thread two of the pipeline. We consider this data to be of two different `modes' and thus require a method of combining the information from both modes. For a classifier we select the Random Forest as it is inherently explainable and has been shown to perform well on EEG classification from features \cite{edla2018classification} \cite{fraiwan2012automated}. For fusion of the two modes we can select feature level fusion, decision level fusion or propose another method. As the number of features per electrode can be far higher than the number of connectivity metrics we decide against direct feature level fusion. For tree-based models (especially Random Forests) this can skew the model towards the mode with more features. Decision-level fusion would be viable. However, we hypothesize that interactions between local electrode and global connectivity features are likely useful for learning. 
We therefore use a balanced sampling strategy for the trees of the Random Forest whereby each tree has a random subset of features selected from mode 1 (per electrode features) and mode 2 (connectivity features). This ensures that each tree and each split within that tree has access to both local electrode features and brain connectivity features. This balanced sampling strategy is detailed in \ref{alg:fusionforest}.

\begin{algorithm}[tb]
\caption{Balanced sampling for Random Forest using two feature modes ($X_a$ and $X_b$), independently select a random subset of features from each (lines 5-6) before fitting each of the $K$ trees. This prevents the feature set with a higher dimension from dominating the model.}

% \caption{Fusion Forest}
\label{alg:fusionforest}
\begin{algorithmic}[1]
    \STATE \textbf{Input:} $N$ samples, $Xa$ with $A$ total features (mode 1), $Xb$ with $B$ total features (mode 2)  
    \STATE Select number of trees $K$ for the Random Forest
    \FOR{$k = 1$ to $K$}
      \STATE Select random subset of samples $N_{\text{subset}}$ (bootstrap sample of size $N$)
      \STATE Select a random subset of $\sqrt{A}$ features from $Xa$ giving $Xa_{\text{subset}}$
      \STATE Select a random subset of $\sqrt{B}$ features from $Xb$ giving $Xb_{\text{subset}}$
      \STATE Fit a Decision Tree to $\bigl(N_{\text{subset}}, Xa_{\text{subset}}, Xb_{\text{subset}}\bigr)$
      \STATE Add the Decision Tree to the Random Forest
    \ENDFOR
\end{algorithmic}
\end{algorithm}

\section{Experiments and Discussion} %Label is to be confirmed

\subsection{Datasets}

We use five publicly available EEG datasets. Three of the datasets recorded data with 14 channel Emotiv EEG headsets. These datasets are DREAMER (emotion detection) \cite{katsigiannis2017dreamer} , STEW (Mental workload Classification)  \cite{lim2018stew} and Crowdsourced (eyes open / close detection) \cite{williams2023crowdsourced}. EEG from these datasets had a sampling rate of 128Hz upon recording and all samples are  two seconds long (after pre-processing). Recorded data for these three datasets is processed as per \cite{mohammadi2024eeg2rep}, snipping the data to create 2 second segments. Further details are provided in \ref{datasets}.

The other two datasets are from the Temple University Hospital (TUH) Corpus \cite{TUAB_dataset} \cite{TUEV_dataset}. TUH is one of the largest EEG data repositories in the world with data collected in a lab setting with a range of different EEG recording devices. In processing this data we select 16 standard EEG channels following the 10-20 international system. This data was recorded at 256Hz with 5 second samples for TUEV \cite{TUEV_dataset} and 10 second samples for TUAB \cite{TUAB_dataset}. We provide further details of processing in Appendix \ref{datasets}. 

The Emotiv datasets were split into train / validation / test sets subject wise, challenging models to learn generalizable patterns across subjects. The TUH datasets (TUEV and TUAB) were inherently split already into train and test sets. We further split the train sets into 80\% training and 20\% validation. Again, this is following the same protocol as in \cite{mohammadi2024eeg2rep}.

Table \ref{eeg_datasets} shows the main characteristics of all five data sets. Further information on each dataset and pre-processing can be found in  \ref{datasets}.

\begin{table}[htbp]
\centering
\caption{Properties of the EEG datasets used (adapted from \cite{mohammadi2024eeg2rep}). DREAMER \cite{katsigiannis2017dreamer}, STEW \cite{lim2018stew}, Crowdsourced \cite{williams2023crowdsourced}, TUEV \cite{TUEV_dataset} and TUAB \cite{TUAB_dataset}  }
\label{eeg_datasets}
\vskip 0.15in
\renewcommand{\arraystretch}{1.5} 
\resizebox{\columnwidth}{!}{%
\begin{tabular}{lclccc}
\toprule
\textbf{Dataset} & \textbf{Classification Task} & \textbf{Dim.} & \textbf{Freq.} & \textbf{Duration} & \textbf{Samples} \\
\midrule
DREAMER      & Emotion Detection  & 14  & 128Hz  & 2s  & 77,910   \\
STEW         & \makecell{Mental Workload}  & 14  & 128Hz  & 2s  & 26,136   \\
Crowdsourced & \makecell{Eyes Open/Closed } & 14  & 128Hz  & 2s  & 12,296   \\
TUEV        & Event Detection & 16  & 256Hz  & 5s  & 112,464  \\
TUAB        & \makecell{Abnormal EEG } & 16  & 256Hz  & 10s & 409,455  \\ 
\bottomrule
\end{tabular}
} 
\renewcommand{\arraystretch}{1} % Reset to default
\end{table}

\subsection{State-of-the-art Methods } \label{competitors}
We compare our pipeline to many state-of-the-art methods from the literature. Recent research has seen representation based deep learning methods excel on EEG data \cite{kostas2021bendr}\cite{mohammadi2024eeg2rep}. Thus we compare versus four deep learning representation based classification models. The recently developed EEG2Rep \cite{mohammadi2024eeg2rep} as well as BIOT \cite{yang2024biot}, BENDR \cite{kostas2021bendr} and MAEEG \cite{chien2022maeeg}. Per \cite{mohammadi2024eeg2rep} these self-supervised models achieve the highest performance if they are first able to learn a representation of the EEG by training on the data in a self-supervised fashion without labels before being trained on the data with labels. This is referred to as the `Fine Tuning' setting. Thus, we deploy these models in this fine-tuning fashion where they perform best. We deploy EEG2Rep in both the fine-tuning fashion and the default fashion where it is deployed directly on the data without a pre-training phase. We also select the EEG Conformer \cite{eegconformer} as a competitor model (deep learning model for classification). This model is adapted from transformer architecture.  Finally, we select a general time series feature-based method, Catch22 \cite{lubba2019catch22}. These models were all implemented using publicly available code to ensure fair evaluation.

\subsection{Experimental Procedure} \label{experimental_procedure}

The implementation of the KnowEEG pipeline as described in \ref{method} follows a strict sequential order ; (1) Thread One feature filtering is performed on the training split to identify informative statistical features, (2) Thread Two connectivity parameter selection is completed using validation set, (3) Random Forest tuning is performed using the validation set, (4) KnowEEG Model is tested on the hold out test set.

For all datasets the feature based pipeline KnowEEG (see \ref{method} ) and all competitor models (see \ref{competitors}) were trained on the training split of the dataset, had hyperparameters tuned using the validation dataset and were tested on a holdout test set. This was done for five different random seeds. Mean and standard deviation of performance metrics were used to assess model performance. For binary classification, accuracy and Area Under ROC Curve (AUROC) are selected as the performance metrics. For multi-class classification on the TUEV dataset Balanced Accuracy and Weighted F1 score are used. Balanced Accuracy assigns equal weight to each class, ensuring the model is penalized for poor performance on minority classes. In contrast, the Weighted F1-score weights class-specific F1-scores by their frequency, thereby providing a performance measure that reflects the model's effectiveness across the dataset without allowing minority classes to exert a disproportionate influence. Together, these metrics provide a comprehensive overview of model performance. This protocol follows the experimental procedure established in \cite{mohammadi2024eeg2rep}.

For KnowEEG the TSFresh \cite{christ2018time} package is used with `Efficient' settings to calculate per electrode statistics. The MNE Python package \cite{MNE_2013} is used for calculating connectivity metrics. For the Random Forest (with balanced sampling) the number of trees hyperparameter was selected from the list [50,100,200,500, 800, 1000]. Further details for the set up of KnowEEG are provided in \ref{appendix_knowEEG}.

\subsection{Results} \label{results}

% Modify this results section to do some statistical analysis on the results, which is better / worse and why
Table \ref{res_table} shows the average performance for all models across the DREAMER, Crowdsourced, STEW, TUEV and TUAB datasets. With 10 total performance metrics across 5 datasets the feature based pipeline KnowEEG achieves best performance vs. all other models on 5/10 metrics, second best on 4/10 and third best on 1/10 metrics. Thus, judging by performance ranking alone, the feature based KnowEEG is competitive with state of the art deep learning models. 

Crowdsourced is the EEG task where models perform best with the task of classifying eyes open vs eyes closed EEG data. On this dataset, KnowEEG  has the second highest accuracy with 92.81\% performing marginally worse than the pre-trained EEG2Rep model. For AUROC KnowEEG outperforms all other models with 96.72 and performs consistently across seeds with a small standard deviation of only 0.21. Notably, the other feature based method Catch22 \cite{lubba2019catch22} performs well on this dataset achieving 89.57 \% accuracy and 95.82 AUROC (second best). However, Catch22 performs poorly on all other datasets illustrating that feature-based methods can struggle to perform well across a diverse set of EEG classification tasks if the feature set is not diverse enough.

On STEW with the task of mental workload classification KnowEEG markedly outperforms all other models in both accuracy and AUROC with standard deviations of under 1 for both metrics again illustrating consistent performance across seeds. On DREAMER performance is more variable across seeds and closer to other models.  

For the TUEV dataset KnowEEG is third in balanced accuracy yet outperforms all other models in weighted F1 score by a substantial margin. TUEV is an extremely unbalanced dataset with 6 classes of EEG events. Therefore in this case balanced accuracy is not necessarily the best measure of overall model performance. This is because balanced accuracy ignores class distributions resulting in smaller classes having a disproportional impact on the balanced accuracy score and can be a drawback if targeting good accuracy on the entire dataset\cite{grandini2020metrics}. Weighted F1-score combines both precision and recall with each class weighted proportionally. Thus, it is notable that the KnowEEG feature based pipeline outperforms all other models in this metric by a large margin.

% For abnormal EEG classification on TUAB KnowEEG performs second best to EEG2Rep on both accuracy and AUROC. However KnowEEG is within the standard deviation of EEG2Rep so this result is not significant. BIOT also reaches the same performances as EEG2Rep and KnowEEG when accounting for standard deviation across runs.

For abnormal EEG classification on TUAB KnowEEG performs second best to EEG2Rep on both accuracy and AUROC, with BIOT third. The three models differ by only 1.31\% in accuracy and 1.01 in AUROC. We therefore consider their performance on this dataset as comparable. 

EEG2Rep Pre-trained \cite{mohammadi2024eeg2rep} is the best performing deep learning model achieving best performance or second best performance on 8/10 metrics across the five datasets. Neither EEG2Rep Pre-trained or KnowEEG significantly outperform each other across the five datasets when using the Wilcoxon Signed-Rank test. Besides performance, when compared with Deep Learning models, a feature based pipeline such as KnowEEG has the advantage of not requiring a Graphics Processing Unit (GPU) for training. KnowEEG also bypasses both the intensive pre-training typical of state-of-the-art self-supervised models and the quadratic computational complexity $O(L^2)$ inherent to transformer architectures during iterative training. For further detail on computational complexity see appendix \ref{appendix:complexity}. All self-supervised representation-based models (except EEG2Rep Default) were trained first on the training data in self-supervised fashion and then tuned in supervised fashion. Finally, KnowEEG also has the advantage of using calculated features that are inherently interpretable and a tree-based model which enables users to access feature importances directly. We demonstrate the benefits of this explainability in Section \ref{analysis}.

\definecolor{customblue}{RGB}{230, 240, 255} % Adjust RGB values as needed

\begin{table}[htbp]
\renewcommand{\arraystretch}{1.5} 
\setlength{\tabcolsep}{5pt}  % Reduce column spacing to fit table better
\centering
\caption{Performance of the feature based KnowEEG pipeline versus competitor models across datasets. KnowEEG results are shaded in blue. Best performance per dataset is in \textbf{bold} with second best \underline{underlined} with Accuracy results in table a) and AUROC in table b). }
\label{res_table}
% ACCURACY TABLE
\begin{subtable}{\linewidth}
    \centering
    \caption{Accuracy (Acc) Performance}
    \label{know_eeg_acc_table}
    \resizebox{\linewidth}{!}{
    \begin{tabular}{|l|c|c|c|c|c|}
    \hline
    \textbf{Models} & \textbf{DREAMER} & \textbf{Crowdsourced} & \textbf{STEW} & \textbf{TUEV} & \textbf{TUAB} \\
    & \textbf{Acc} & \textbf{Acc} & \textbf{Acc} & \textbf{B-Acc} & \textbf{Acc} \\
    \hline
    BENDR  & 54.45$\pm$2.11  & 83.78$\pm$2.35  & 69.74$\pm$2.11  & 41.17$\pm$2.89  & 76.96$\pm$3.98 \\
    MAEEG  & 53.63$\pm$2.61  & 86.75$\pm$3.50  & 72.46$\pm$3.67  & 41.23$\pm$3.65  & 77.56$\pm$3.56 \\
    BIOT  & 53.45$\pm$2.01  & 87.95$\pm$3.52  & 69.88$\pm$2.15  & \underline{46.02}$\pm$1.68  & 79.21$\pm$2.15 \\
    Catch22 & 59.06$\pm$1.61  & 89.57$\pm$0.64  & 55.72$\pm$0.2  & 33.65$\pm$0.5  & 69.84$\pm$1.06 \\
    EEGConformer  & 55.40$\pm$2.76  & 87.62$\pm$2.91  & \underline{74.35}$\pm$1.88  & 43.88$\pm$1.62  & 78.36$\pm$1.71 \\
    EEG2Rep (Default) & 54.61$\pm$2.22  & 91.19$\pm$1.18  & 70.26$\pm$1.59  & 44.25$\pm$3.01  & 77.85$\pm$3.14 \\
    EEG2Rep (Pre-Trained) & \underline{60.37}$\pm$1.52  & \textbf{94.13}$\pm$1.21  & 73.60$\pm$1.47  & \textbf{52.95}$\pm$1.58  & \textbf{80.52}$\pm$2.22 \\
    \rowcolor{customblue} \textbf{KnowEEG} & \textbf{61.53}$\pm$2.59  & \underline{92.81}$\pm$0.32  & \textbf{77.86}$\pm$0.19  & 45.06$\pm$0.71  & \underline{80.18}$\pm$0.24 \\
    \hline
    \end{tabular}
    } % end resizebox
\end{subtable}

\vspace{0.5cm} % Adds spacing between the two tables

% AUROC TABLE
\begin{subtable}{\linewidth}
    \centering
    \caption{AUROC Performance}
    \label{know_eeg_auroc_table}
    \resizebox{\linewidth}{!}{
    \begin{tabular}{|l|c|c|c|c|c|}
    \hline
    \textbf{Models} & \textbf{DREAMER} & \textbf{Crowdsourced} & \textbf{STEW} & \textbf{TUEV} & \textbf{TUAB} \\
    & \textbf{AUROC} & \textbf{AUROC} & \textbf{AUROC} & \textbf{W-F1} & \textbf{AUROC} \\
    \hline
    BENDR  & 53.02$\pm$1.31  & 83.80$\pm$2.63  & 69.77$\pm$2.03  & 67.31$\pm$2.96  & 83.97$\pm$3.44 \\
    MAEEG  & 52.08$\pm$2.36  & 86.21$\pm$3.41  & 72.50$\pm$3.22  & 67.38$\pm$3.69  & 86.56$\pm$3.33 \\
    BIOT  & 53.53$\pm$1.82  & 87.78$\pm$3.09  & 70.11$\pm$2.57  & 69.98$\pm$1.99  & 87.42$\pm$2.01 \\
    Catch22 & 51.6$\pm$2.17  & \underline{95.82}$\pm$0.3  & 53.2$\pm$0.14  & 67.01$\pm$0.34  & 66.16$\pm$1.12 \\
    EEGConformer  & 54.19$\pm$2.57  & 87.24$\pm$2.06  & 73.11$\pm$1.95  & 67.20$\pm$2.31  & 83.89$\pm$2.64 \\
    EEG2Rep (Default) & 53.61$\pm$2.07  & 91.22$\pm$1.23  & 69.77$\pm$2.03  & 68.95$\pm$2.89  & 84.91$\pm$3.07 \\
    EEG2Rep (Pre-Trained) & \textbf{59.42}$\pm$1.45  & 94.13$\pm$2.17  & \underline{74.40}$\pm$1.50  & \underline{75.08}$\pm$1.21  & \textbf{88.43}$\pm$3.09 \\
    \rowcolor{customblue} \textbf{KnowEEG} & \underline{55.29}$\pm$2.58  & \textbf{96.72}$\pm$0.21  & \textbf{85.96}$\pm$0.8  & \textbf{78.28}$\pm$0.56  & \underline{87.71}$\pm$0.23 \\
    \hline
    \end{tabular}
    } % end resizebox
\end{subtable}
\end{table}

\subsection{Feature Based KnowEEG Explainability Analysis} \label{analysis}

Here, we analyse the KnowEEG model trained on the training and validation sets for the Crowdsource dataset and demonstrate the key explainability benefits. The Crowdsource task presents the problem of binary classification on 14-channel EEG data with the two classes being eyes open and eyes closed. This task is selected for explainability analysis because the eyes open versus eyes closed states are well studied in neuroscientific literature allowing for comparison between the models explanations and established ground truths. Furthermore, explainability analysis is most informative on tasks where the model performs well as analysis of poor predictions would provide negligible insight into discriminative class features. KnowEEG performs best on the Crowdsource dataset with AUROC 96.72 and an accuracy of 92.81\%. For interpretability analysis, we propose analysing the connectivity and statistical features separately.

In hyperparameter selection, Functional Power Connectivity (FPC) was selected as the connectivity measure for the Crowdsource dataset. As defined in \ref{FPC}, the FPC feature set is calculated between each electrode pair over six defined frequency bands (alpha, theta, delta, sigma, beta, gamma) as well as having additional features indicating relative band power values per electrode. The FPC features therefore describe both band power activity and functional connectivity across the brain.

\begin{figure}[h!]
        \centering
        \includegraphics[width=1.0\columnwidth]{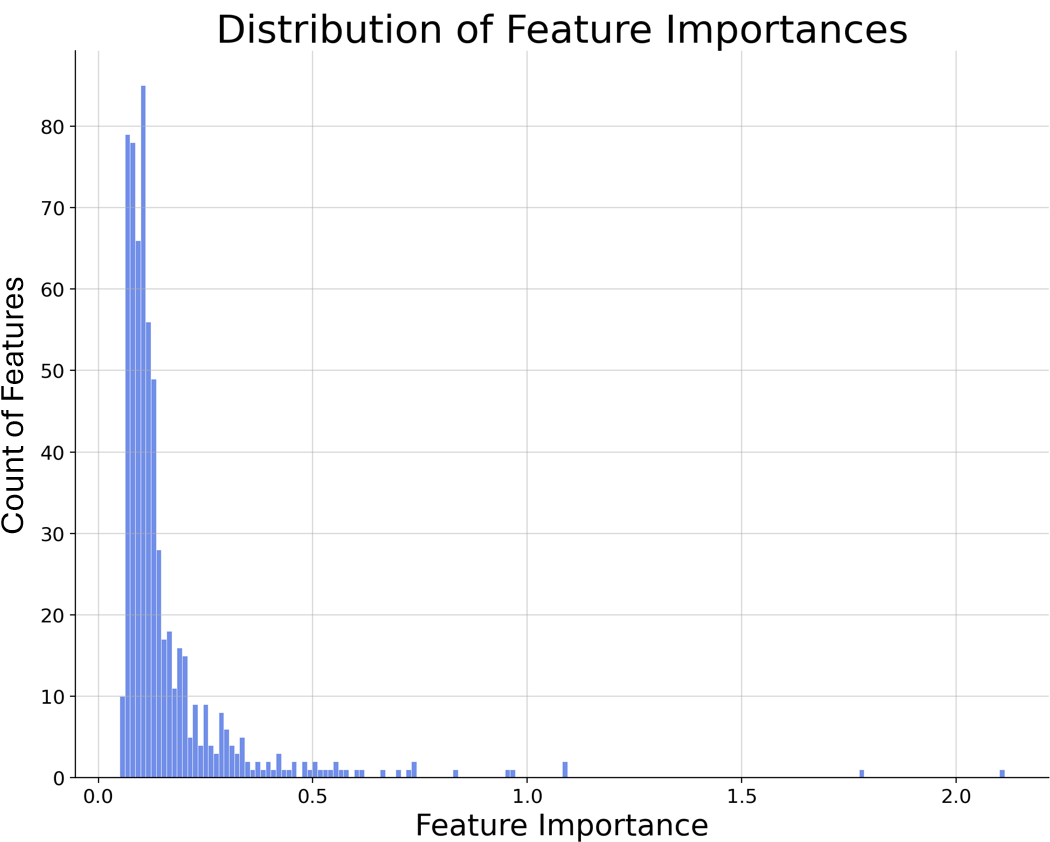}
        \caption{This histogram displays the Gini importance values for the Functional Power Connectivity features (FPC) used by the trained KnowEEG model on the Crowdsource dataset. The x-axis represents the feature importance score with the y-axis showing the count of features at that importance level. The distribution is highly skewed with an exponential-like decay, showing that a large majority of features have an importance score near zero, while a small group of features have higher importance scores.}
        \label{con_importances}
\end{figure}

A first step in analysing these feature importances is to plot their distribution. Figure \ref{con_importances} shows an exponential like feature distribution with many feature importances with low values close to zero and fewer and fewer features with higher importances. 

\begin{figure}[h!]
        \centering
        \includegraphics[width=0.9\columnwidth]{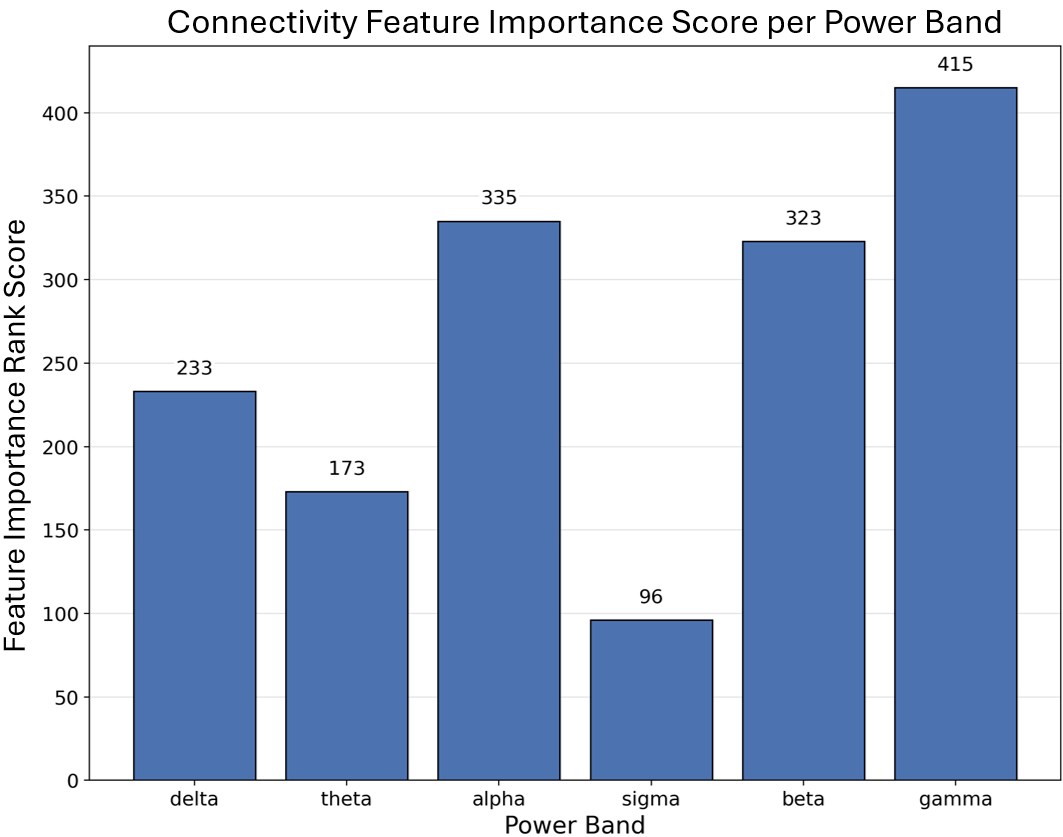}
        \caption{Feature Importance score per power band for Functional Power Connectivity on the Crowdsource dataset. FPC gamma features have the highest feature importance rank score overall followed by alpha, beta, delta, theta and then sigma with the lowest score.}
        \label{con_band_importances}
\end{figure}

We propose as a simple next step to analyse connectivity feature importances per power band. Due to interactions between features, summing feature importances per band does not accurately represent group feature importance \cite{NEURIPS2020_c7bf0b7c}. Therefore, instead we compare each channel-channel connectivity measure per band and rank them. For example, electrodes AF3-AF4 will have 6 FPC connectivity values, one for each band. The band with the highest feature importance for each electrode-electrode pair receives a score of 5 down to 6th place which receives a score of 0. The per band ranking feature importance scores are shown in figure \ref{con_band_importances}. 

% FPC gamma features have the highest feature importance rank score overall followed by alpha, delta, theta and sigma with the lowest score.

% Connectivity Explainability Starts Here _____________________________________________________________________________

\begin{figure}[h!]
        \centering
        \includegraphics[width=0.9\columnwidth]{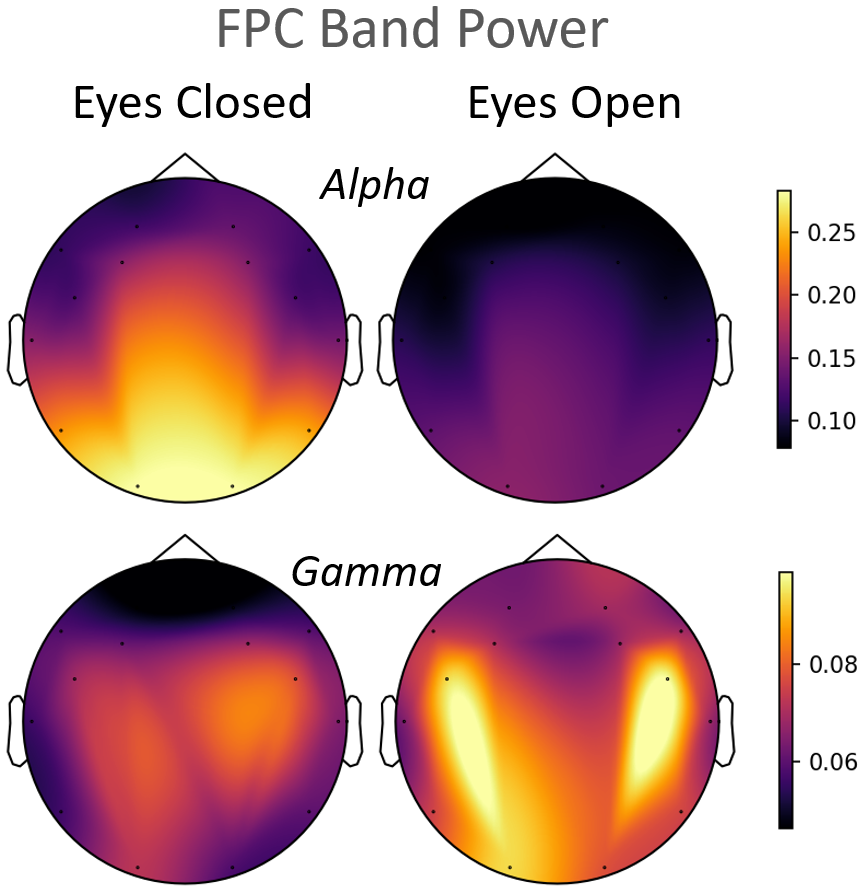}
        \caption{This plot shows mean relative band power in the alpha and gamma bands for eyes closed (left) and eyes open (right) participants. For the Alpha band the eyes closed group have a higher mean power particular in the Occipital Lobes towards the back of the head. For the Gamma band the eyes open group have a higher mean power with peaks in the Left and Right Temporal / Central regions of the brain.}
        \label{FPC_Band_Power_Fig}
\end{figure}

\begin{figure}[h!]
        \centering
        \includegraphics[width=0.9\columnwidth]{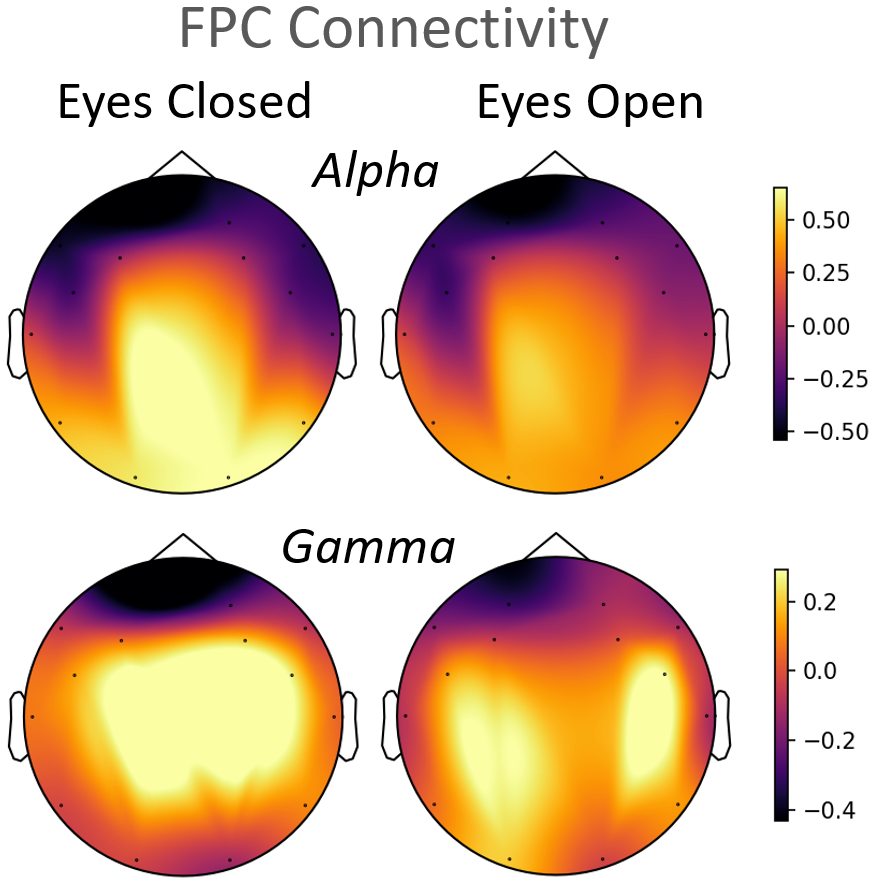}
        \caption{This plot shows the mean Alpha and Gamma band connectivity for the eyes closed (left) and eyes open (right) groups per the FPC features (defined in \ref{FPC}). The eyes closed group have higher Alpha connectivity across the brain particularly in the Occipital regions. For Gamma connectivity the differences between the groups are not visually significant.}
        \label{FPC_Connectivity_Fig}
\end{figure}

Per figure \ref{con_band_importances}, the Functional Power Connectivity features on the gamma band have the highest importance followed by alpha, beta, delta, theta and then sigma with the lowest importance rank. This suggests that FPC features in gamma and alpha bands are most discriminative versus other bands for eyes closed versus eyes open classification. We can take this analysis further by visualizing FPC features for alpha and gamma for the two classes. We can do this on a surface plot of the 14 electrodes on the head, with the FPC band power features separated from FPC pure connectivity features. We plot FPC band power in figure \ref{FPC_Band_Power_Fig} and connectivity in figure \ref{FPC_Connectivity_Fig}.

Per figure \ref{FPC_Band_Power_Fig} the mean eyes open alpha band power is noticeably higher in the eyes closed group vs. eyes open, particularly in the Occipital electrodes towards the back of the head. This result is in line with the neuroscience literature that show an increase in alpha power in the eyes closed state \cite{gomezram_eyes_op_close, FPC_explainability_4}. Contrastively the mean gamma band power is significantly higher across the head for the eyes open state vs. eyes closed per figure \ref{FPC_Band_Power_Fig}. Again, this is an expected result supported by EEG studies \cite{FPC_explainability_3}. The eyes being open increases alertness and raises arousal in the brain resulting in stronger activity in the gamma band. 

Figure \ref{FPC_Connectivity_Fig} shows that mean FPC alpha connectivity is higher for the eyes closed participants vs. eyes open. This phenomenon is known as alpha de-synchronization and is consistent with EEG literature that shows "functional connectivity in the alpha band decreases in the eyes open condition compared to eyes closed" \cite{gomezram_eyes_op_close}. Mean FPC gamma connectivity does not show significant differences between the eyes open and closed groups. This likely reflects the low amplitude and high variability of resting-state gamma activity, making gamma band connectivity a weak discriminator for distinguishing between the eyes open and closed states.

In summary, from our analysis of the Functional Power Connectivity features we have uncovered known neuroscience knowledge. Specifically, we have found increased mean alpha band activity and connectivity in the eyes closed state compared to eyes open as well as increased mean gamma band activity in the eyes open class \cite{gomezram_eyes_op_close, FPC_explainability_1, FPC_explainability_3, FPC_explainability_4}. This ability of feature based pipelines such as KnowEEG to uncover knowledge about the classes via interpretable features is a significant strength. We must note though, that the summary plots showing mean values across the head for both classes will not always show a clear distinction between classes. However, they are a useful summary view and the detailed feature importance scores will always be available to guide more detailed analysis of the classes.

Furthermore, a granular per channel analysis could also be completed to determine if the connectivity between any channel pairs is particularly discriminative for the classification problem. To keep the analysis focused we move on to the per channel statistical features.

% Connectivity Feat. Importance End___________________________________________________________________

The feature importances for per channel statistics also follow an exponential-like distribution with many features having small importances close to 0 and fewer features having higher importances. As a simple first step we can analyse the top n features by feature importance. We select n = 10 features presenting these features in Table \ref{tab:feature_importance} below. 

\begin{table}[htbp]
\centering
\caption{Statistical features with highest importance ranked 1 to 10.  The table displays the feature name, parameters specifying feature calculation, corresponding electrode and corresponding brain region. It is notable that the top 7 features are all from the Occipital brain region.}
\renewcommand{\arraystretch}{1.3} 
\label{tab:feature_importance}
\vskip 0.1in
\resizebox{\columnwidth}{!}{%
\begin{tabular}{c l l l l}
    \hline
    \textbf{Rank} & \textbf{Feature} & \textbf{Parameters} & \textbf{Channel} & \textbf{Brain Region}\\
    \hline
     1  & Permutation Entropy    & Dimension 6 Tau 1     & O1  & Occipital \\
     2  & Permutation Entropy    & Dimension 4 Tau 1     & O2  & Occipital \\
     3  & Permutation Entropy    & Dimension 5 Tau 1     & O2  & Occipital \\
     4  & Partial Autocorrelation & Lag 2                & O1  & Occipital \\
     5  & Number Peaks           & Support 1             & O2  & Occipital \\
     6  & Permutation Entropy    & Dimension 5 Tau 1     & O1  & Occipital \\
     7  & Permutation Entropy    & Dimension 3 Tau 1     & O2  & Occipital \\
     8  & Mean Absolute Change   & N/A                   & FC6 & Right Central \\
     9  & Permutation Entropy    & Dimension 7 Tau 1     & O1  & Occipital \\
     10 & Quantile Change Mean   & Upper 0.8, Lower 0    & FC6 & Right Central \\
    \hline
\end{tabular}%
}
\renewcommand{\arraystretch}{1} % reset to default
\end{table}

From Table 3 it is notable that 7 out of the top 7 highest importance features are from the Occipital brain region. This suggests that the Occipital brain region provides features that are most discriminative of the two classes. The two classes are eyes open and eyes closed. We know from existing neuroscience research that the Occipital region of the brain is responsible for visual processing as it "houses the visual cortex responsible for processing and interpreting visual stimuli" \cite{fossa2024occipital}. Therefore, again, from the KnowEEG feature based pipeline we have been able to discover knowledge (signals from the Occipital brain region being most discriminative for the two classes) and have verified this versus existing neuroscience literature. This shows that our pipeline can aid discovery of new information where differences between classes are not yet understood.

\begin{figure}[h!]
        \centering
        \includegraphics[width=1.0\columnwidth]{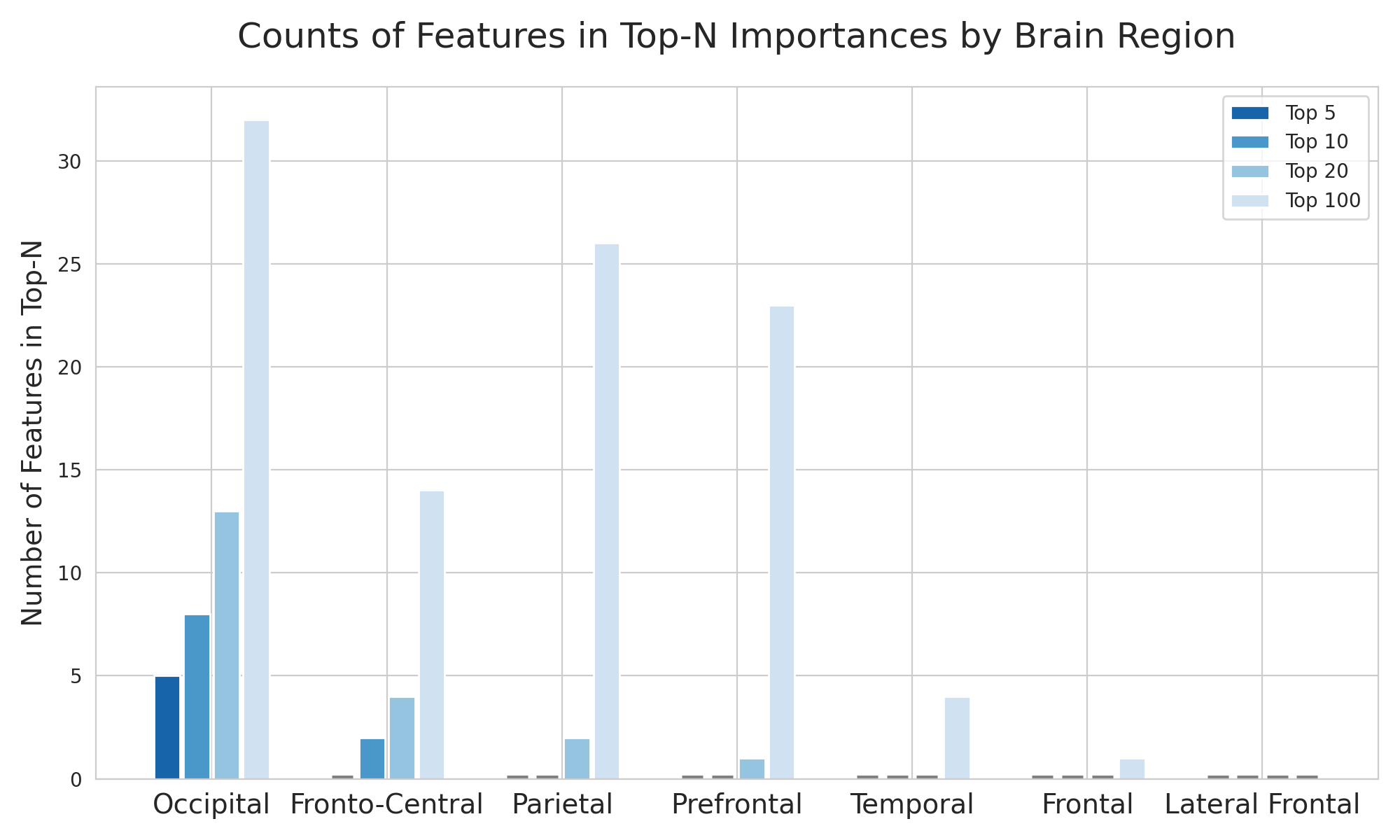}
        \caption{Count of features in top N importances (N = 5, 10, 20, 100) by Brain Region. From this bar chart it is clear that the Occipital region has a disproportionately large number of highly discriminative features vs. other regions. The chart shows that 5 out of the top 5 (dark blue) most important features are from the Occipital region as well as 7 of the top 10, 13 of the top 20 and 32 of the top 100.}
        \label{feat_import_graph}
\end{figure}

We extend this feature importance ranking beyond the top 10 features. In Figure \ref{feat_import_graph} we count how many features from each brain region are within the top 5, 10, 20 and 100 features by feature importance. The Occipital brain regions continues to outscore all other regions in the Top 20 and Top 100 features confirming this finding. 

Another interesting finding from Table \ref{tab:feature_importance} is that the three highest importance features and 6 of the top 10 are permutation entropy from electrodes in the Occipital brain region. Permutation entropy is considered as `a natural complexity measure for time series' \cite{bandt2002permutation}. The calculation of permutation entropy requires two parameters (Dimension and Tau) which is why this feature can appear multiple times for the same electrode in Table \ref{tab:feature_importance}.

\begin{figure}[h!]
        \centering
        \includegraphics[width=1.0\columnwidth]{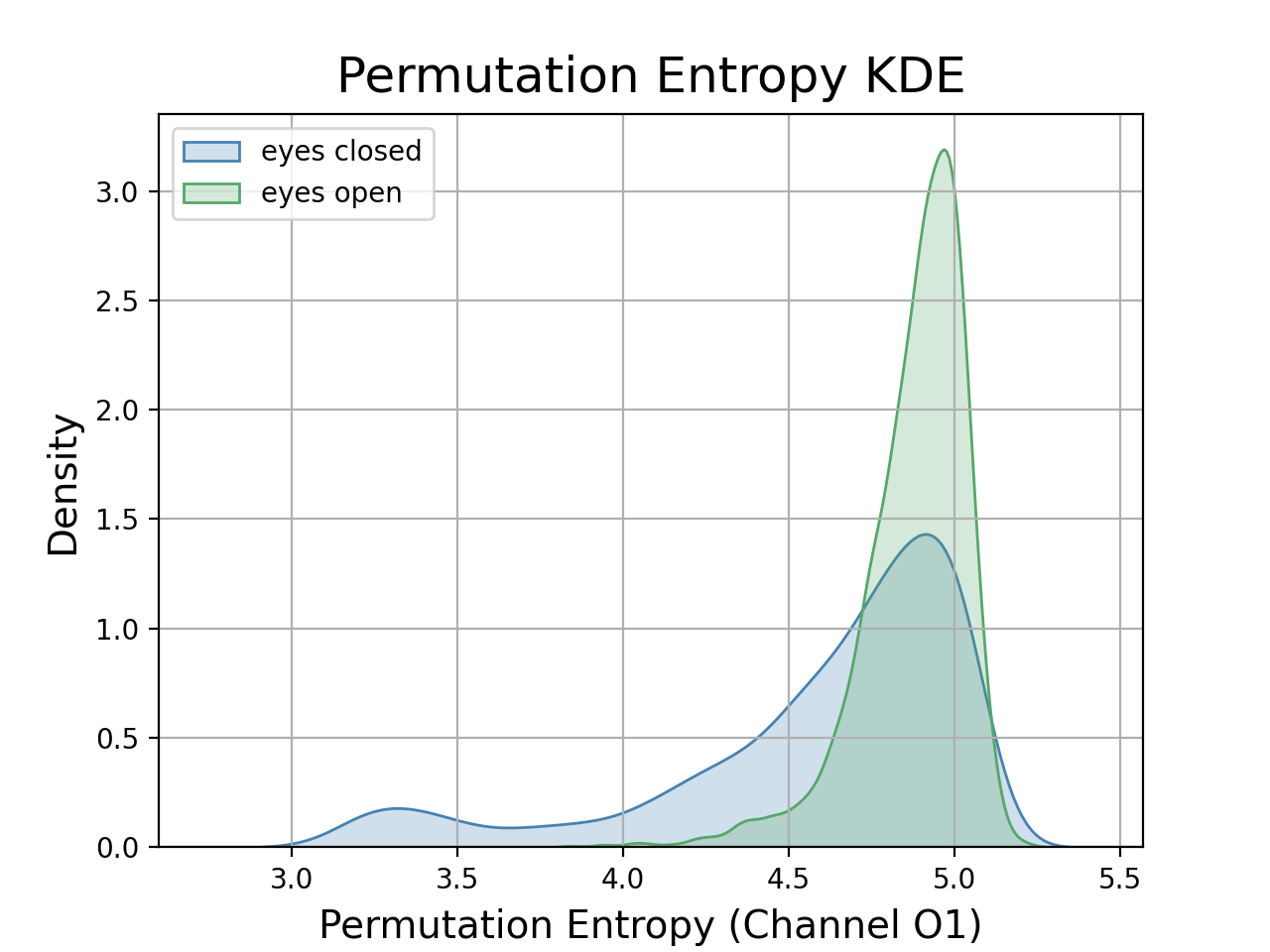}
        \caption{KDE plot of the rank 1 feature from Table \ref{tab:feature_importance}: Permutation Entropy of the O1 (Occipital Lobe) EEG channel. The distribution for the `eyes open' class (green) is distinctly narrower and has a higher peak density, while the `eyes closed' class (blue) shows a broader distribution with a wider spread of values.}
        \label{perm_entropy_kde}
\end{figure}

We plot the Kernel Density Estimates for eyes closed and eyes open for the highest importance ranked feature in Figure \ref{perm_entropy_kde}. It is clear from the figures that the eyes closed group has a much broader distribution compared with the eyes open group. The eyes open group distribution is narrower (standard deviation 0.16 vs 0.47) with a slightly higher peak. When testing the distributions for this feature, we use the two sample Kolmogorov–Smirnov test and find that the two distributions vary significantly (p-value\textless0.0001). 

Therefore, again our pipeline has enabled us to find out useful information about our classes. In this case that the distributions for permutation entropy on the electrode O1 channel calculated with dimension 6 and tau 1 are significantly different for the eyes closed and eyes open groups.

\subsection{Benefits and Limitations} \label{Benefits}
\paragraph{Benefits}
The main benefit of feature based pipelines such as KnowEEG is the explainability of the model which is in contrast to many state of the art deep learning models. The features themselves have meaning in the generalized time series or EEG domain and the Random Forest as a tree-based model allows users to directly access feature importances. We demonstrate this benefit in \ref{analysis}. We show via the Crowdsourced dataset that this can allow users to discover physiologically meaningful biomarkers for identifying the two classes. Notably, we find Occipital Lobe signal statistics aswell as gamma band activity, alpha band activity and alpha band connectivity to be discriminative of the eyes open and eyes closed classes. Both findings are confirmed by existing neuroscience literature.  Another benefit of a feature based pipeline that uses a broad set of diverse features is high performance across different EEG classification tasks. Versus 7 state-of-the-art baselines on five datasets on 10 performance metrics the feature based KnowEEG pipeline is best on 5/10 metrics, second on 4/10 and outside of the top two models on only 1/10 metrics. Overall, the feature based KnowEEG pipeline is competitive versus the best performing deep learning models across 5 different EEG datasets. 

Finally, our pipeline has the significant benefit of not requiring GPU resources for training and a lower computational cost overall versus state of the deep learning models. KnowEEG does not require the intensive pre-training typical of state-of-the-art self-supervised models nor does it suffer the quadratic computational complexity $O(L^2)$ inherent to transformer architectures during iterative training. Once deterministic feature calculation is complete, KnowEEG memory and training requirements remain independent of sequence length, significantly lowering the hardware barrier compared to deep learning baselines. For further detail on computational complexity analysis see appendix \ref{appendix:complexity}.

\paragraph{Limitations} \label{Limitations}
One limitation is regarding the explainability of feature based pipelines such as KnowEEG. The explanations are only useful if the model itself performs well. If the model performs poorly then the feature importances are likely to provide little insight into the classes. Furthermore for tree-based models importance measures such as the Gini Index can be biased. Gini index is known to be biased towards high cardinality features that offer more split points and can also suffer from correlation dilution where the importance of highly correlated features is split between each feature \cite{strobl2008conditionalcorrelation} \cite{strobl2007biashighcardinality}. To address these biases, feature analysis should be done with regard to clusters of features and regions e.g. Occipital or band powers with regards to connectivity as per \ref{analysis}. This analysis is more robust to permutations in individual feature importances verified by the fact that our findings align with current neuroscience literature. Additionally, one can also use model agnostic explainability tools such as SHAP or LIME to verify analyses of feature importances \cite{SHAPlundberg2017unified} \cite{LIMEribeiro2016should}.  

There also exists limitations in the selected connectivity metrics as they involve inherent trade-offs in the capture of underlying connectivity, and physiological interpretations are strictly dependent on the chosen metric. Phase-based metrics, such as Phase Lag Index and Coherence, uncover temporal synchronization or phase-locking, reflecting the rhythmic firing alignment of neuronal populations across regions \cite{chiarion2023connectivity}. A primary limitation of these metrics is their statistical instability and sensitivity to noise in short-duration signals, a challenge compounded by the requirement for signal segmentation into epochs to ensure reliable estimation \cite{MNE_2013}. In contrast, Functional Power Connectivity (FPC) captures connectivity through the spatial distribution and relative differences in oscillatory power across regions. While FPC offers greater stability for short signals because it bypasses the need for epoch-based segmentation, it inherently discards temporal phase information. These distinct mechanisms must be considered when evaluating the neurophysiological basis of the model’s features.

% Add additional paragraph here in response to the viewers point around pre-processing choices on the EEG data.
Additionally, our preprocessing choices including window lengths, segment durations, and frequency band definitions follow established protocols from existing literature and the EEG2Rep framework to ensure a fair benchmark \cite{mohammadi2024eeg2rep}. However, the sensitivity of the KnowEEG pipeline to changes in these parameters has not been deeply explored. Longer-duration signals, such as the 10-second samples in the TUAB dataset, leave more room for trying alternative epoch segmentation lengths. These initial setup choices directly impact the final feature set. Therefore, evaluating how different preprocessing configurations influence classification performance remains an important area for future sensitivity analyses and could impact KnowEEG performance.

The KnowEEG architecture balances interpretability and computational efficiency through specific design choices. For instance, while univariate feature filtering effectively manages high-dimensional feature spaces, it risks discarding complex multivariate interactions. Furthermore the connectivity metric is selected separately from the statistical features meaning that interactions between the connectivity features and statistical features are not considered during the filtering stage. This is helpful for the computational efficiency of the pipeline but again does risk discarding meaningful interactions. Secondly for interpretability the KnowEEG pipeline selects a single connectivity metric. However, allowing for multiple connectivity metrics could increase representational richness and potentially lead to improved performance. Additionally, choosing the optimal metric via a validation set poses a theoretical risk of bias; however, our subject-wise splitting strategy mitigates this by requiring generalization across individuals. Ultimately, these limitations arise from pipeline design choices with the current model balancing interpretability and computational efficiency whilst achieving performance competitive with state-of-the-art deep learning models.

\subsection{Future Work} \label{Future_Work}
% Extend to new datasets  /  
% I think this future work needs to better position this KnowEEG versus current work

Future work will explore additional EEG datasets to determine if the performance of feature based pipelines such as KnowEEG remains high. This should involve Brain Computer Interface EEG datasets such as those discussed in \cite{bci}. BCI datasets broadly address the task of enabling a person's brain signals to communicate with external devices which is distinctly different from the five datasets already explored. 

Future research will also evaluate the clinical utility of KnowEEG for neurodegenerative disease classification, specifically targeting Alzheimer’s and Parkinson’s Disease. Data partitioning into training, validation, and test sets will be executed strictly on a subject-wise basis to ensure the model captures generalized neurophysiological patterns rather than over-fitting to subject-specific artifacts or signal noise. Performance metrics will extend beyond binary diagnostic classification to include the regression of clinical severity scales, such as the Unified Parkinson’s Disease Rating Scale (MDS-UPDRS) \cite{MDS-UPDRS}. The pipeline’s inherent interpretability will then be utilized to determine the most important features for classification. High-ranking features become candidate bio markers and can be assessed versus the established pathophysiology of the disease. Finally, these candidate biomarkers will be evaluated on independent datasets to assess cross-dataset generalization. This methodology provides a systematic framework for the discovery and validation of candidate EEG biomarkers via KnowEEG in clinical settings.

Future work could also look to fuse deep learning and feature based approaches. Deep learning models are very flexible and excel at learning in high data environments whereas feature based models are heavily dependant on the quality of the input features. An example combined approach could involve fusion of pre-calculated features with the latent features of a deep neural network, potentially balancing the strengths of both methods.

Future work could also look to address some of the limitations discussed in \ref{Limitations}. For example one could explore whether allowing multiple connectivity metrics to be selected at once improves performance and whether this performance increase is task specific. Future work could also explore multivariate feature filtering and a joint selection of the statistical features and connectivity metric. This would ensure that multivariate interactions are accounted for during the filtering stage. Though, computational cost would increase an analysis of trade-off between performance increase (if at all) and computational requirements across different EEG tasks could lead to a more optimal model.

Further analysis in future work should be also be done on why the feature based KnowEEG outperforms competitors. For example, one could determine what the extent of the shared information between learnt self-supervised representations such as in EEG2Rep \cite{mohammadi2024eeg2rep} and the pre-calculated EEG features in KnowEEG. This could aid in the development of higher performance models in the future.

% KnowEEG should be deployed on real world data in the healthcare domain to determine if the pipeline can successfully discriminate between classes where the differences between the classes are not already understood. Neurodegenerative diseases such as Alzheimers and Parkinson's would be a good use case. KnowEEG could potentially uncover previously unknown difference in EEG activity between healthy and diseased individuals furthering advancements in this field.

% A feature based pipeline such as KnowEEG could also be deployed on real world healthcare data. Specifically, neurodegenerative diseases such as Alzheimer's and Parkinson's would be a good use case. The data set would be subject wise stratified into train, validation and test data to ensure the model is learning inter-subject patterns. KnowEEG could then be trained to predict the binary presence of Parkinson's or a severity score such as UPDRS. If the model achieves a high classification performance it can then be tested on additional  Parkinson's datasets to determine inter-dataset generalisability. If the model performs on both inter subject and inter dataset generalisability tests then one can use explainability to get the biomarkers. Biomarkers have now been discovered and can be compared vs. existing ones and neurophysiology of these biomarkers can be explored.  

\section{Conclusion} \label{conclusion}

This paper demonstrates that structured feature-based pipelines are a competitive and explainable alternative to deep learning for EEG classification. By integrating signal statistics with brain connectivity metrics through a balanced sampling strategy, the KnowEEG pipeline matches or exceeds state-of-the-art performance across five distinct task domains. These results are achieved on standard CPU hardware, bypassing the high computational and hardware costs associated with complex deep architectures.

The primary advantage of this approach is its inherent interpretability, providing direct access to feature importance rankings. Section \ref{analysis} illustrates the explainability properties of KnowEEG using the Crowdsourced dataset. We were able to correctly identify established differences between the two classes; increased alpha synchronisation, elevated alpha-band power, and lower gamma activity during eyes-closed versus eyes-open with statistics from the Occipital-region electrodes most discriminative for classification. Verifying the model's internal logic against neuroscientific ground truths shows the validity of the KnowEEG explanations.

Future work should evaluate the utility of KnowEEG for biomarker identification across a broader range of datasets, such as motor imagery for brain-computer interfaces and various clinical diagnostic tasks. Furthermore, the shared information between features for KnowEEG and latent spaces of deep neural networks should be studied. This could facilitate the development of hybrid models that combine hand-crafted features with the ability of deep neural networks to learn features directly from raw EEG data.

\section{Funding}
This work was supported by UK Research and Innovation grant EP/\allowbreak S022937\allowbreak/\allowbreak1: Interactive Artificial Intelligence.

\bibliographystyle{elsarticle-num} 
\bibliography{references}

@article{zhang2026stand,
  title={STAND-Net: A Spiking Temporal Attention autoeNcoDer Network for Efficient EEG Artifact Removal},
  author={Zhang, Ruizhi and Guo, Xiaoyu and Pan, Yu and Gao, Shuo},
  journal={IEEE Journal of Biomedical and Health Informatics},
  year={2026},
  publisher={IEEE}
}

@article{ye2024adaptive,
  title={Adaptive spatial--temporal aware graph learning for EEG-based emotion recognition},
  author={Ye, Weishan and Wang, Jiyuan and Chen, Lin and Dai, Lifei and Sun, Zhe and Liang, Zhen},
  journal={Cyborg and Bionic Systems},
  volume={5},
  pages={0088},
  year={2024},
  publisher={AAAS}
}

@article{chen2025novel,
  title={A novel transfer learning-based hybrid EEG-fNIRS brain-computer interface for intracerebral hemorrhage rehabilitation},
  author={Chen, Danyang and Shi, Jian and Tao, Bo and Zhao, Xingwei and Zhao, Zhixian and Li, Shengjie and Xu, Yeguang and Ding, Tao and Zhang, Peng and Ye, Qing and others},
  journal={Advanced Science},
  volume={12},
  number={43},
  pages={e05426},
  year={2025},
  publisher={Wiley Online Library}
}

@article{yue2025augmenting,
  title={Augmenting electroencephalogram transformer for steady-state visually evoked potential-based brain--computer interfaces},
  author={Yue, Jin and Xiao, Xiaolin and Wang, Kun and Yi, Weibo and Jung, Tzyy-Ping and Xu, Minpeng and Ming, Dong},
  journal={Cyborg and Bionic Systems},
  volume={6},
  pages={0379},
  year={2025},
  publisher={AAAS}
}

@article{chen2026brain,
  title={Brain power mapping with optimized deep learning for EEG-based pilot fatigue detection},
  author={Chen, Lerui and Wen, Shengjun and Tang, YukMing and Ma, Yidan and Wang, Haiquan and Geda, Mohammed Woyeso},
  journal={Biomedical Signal Processing and Control},
  volume={114},
  pages={109284},
  year={2026},
  publisher={Elsevier}
}

@article{yu_et_al_2024,
  title={A deep graph kernel-based time series classification algorithm},
  author={Yu, Mengping and Huang, Huan and Hou, Rui and Ma, Xiaoxuan and Yuan, Shuai},
  journal={Pattern Analysis and Applications},
  volume={27},
  number={3},
  pages={73},
  year={2024},
  publisher={Springer}
}

@article{shi2024low_target_detect,
  title={Low-quality video target detection based on EEG signal using eye movement alignment},
  author={Shi, Jianting and Bi, Luzheng and Xu, Xinbo and Feleke, Aberham Genetu and Fei, Weijie},
  journal={Cyborg and Bionic Systems},
  volume={5},
  pages={0121},
  year={2024},
  publisher={AAAS}
}

@article{MDS-UPDRS,
author = {Goetz, Christopher G. and Tilley, Barbara C. and Shaftman, Stephanie R. and Stebbins, Glenn T. and Fahn, Stanley and Martinez-Martin, Pablo and Poewe, Werner and Sampaio, Cristina and Stern, Matthew B. and Dodel, Richard and Dubois, Bruno and Holloway, Robert and Jankovic, Joseph and Kulisevsky, Jaime and Lang, Anthony E. and Lees, Andrew and Leurgans, Sue and LeWitt, Peter A. and Nyenhuis, David and Olanow, C. Warren and Rascol, Olivier and Schrag, Anette and Teresi, Jeanne A. and van Hilten, Jacobus J. and LaPelle, Nancy},
title = {Movement Disorder Society-sponsored revision of the Unified Parkinson's Disease Rating Scale (MDS-UPDRS): Scale presentation and clinimetric testing results},
journal = {Movement Disorders},
volume = {23},
number = {15},
pages = {2129-2170},
doi = {10.1002/mds.22340},
year = {2008}
}

@article{SHAPlundberg2017unified,
  title={A unified approach to interpreting model predictions},
  author={Lundberg, Scott M and Lee, Su-In},
  journal={Advances in neural information processing systems},
  volume={30},
  year={2017}
}

@inproceedings{LIMEribeiro2016should,
  title={" Why should i trust you?" Explaining the predictions of any classifier},
  author={Ribeiro, Marco Tulio and Singh, Sameer and Guestrin, Carlos},
  booktitle={Proceedings of the 22nd ACM SIGKDD international conference on knowledge discovery and data mining},
  pages={1135--1144},
  year={2016}
}

@article{strobl2007biashighcardinality,
  title={Bias in random forest variable importance measures: Illustrations, sources and a solution},
  author={Strobl, Carolin and Boulesteix, Anne-Laure and Zeileis, Achim and Hothorn, Torsten},
  journal={BMC bioinformatics},
  volume={8},
  number={1},
  pages={25},
  year={2007},
  publisher={Springer}
}

@article{strobl2008conditionalcorrelation,
  title={Conditional variable importance for random forests},
  author={Strobl, Carolin and Boulesteix, Anne-Laure and Kneib, Thomas and Augustin, Thomas and Zeileis, Achim},
  journal={BMC bioinformatics},
  volume={9},
  number={1},
  pages={307},
  year={2008},
  publisher={Springer}
}

@article{medical_image_super_resolution,
title = {Analysis of medical images super-resolution via a wavelet pyramid recursive neural network constrained by wavelet energy entropy},
journal = {Neural Networks},
volume = {178},
pages = {106460},
year = {2024},
issn = {0893-6080},
doi = {https://doi.org/10.1016/j.neunet.2024.106460},
author = {Yue Yu and Kun She and Kaibo Shi and Xiao Cai and Oh-Min Kwon and YengChai Soh}
}

@article{noor2025survey_deep_learning,
  title={A survey on state-of-the-art deep learning applications and challenges},
  author={Noor, Mohd Halim Mohd and Ige, Ayokunle Olalekan},
  journal={Engineering Applications of Artificial Intelligence},
  volume={159},
  pages={111225},
  year={2025},
  publisher={Elsevier}
}

@article{modma_dataset,
  title={A multi-modal open dataset for mental-disorder analysis},
  author={Cai, Hanshu and Yuan, Zhenqin and Gao, Yiwen and Sun, Shuting and Li, Na and Tian, Fuze and Xiao, Han and Li, Jianxiu and Yang, Zhengwu and Li, Xiaowei and others},
  journal={Scientific data},
  volume={9},
  number={1},
  pages={178},
  year={2022},
  publisher={Nature Publishing Group UK London}
}

@article{1_recommended_khan2025transfer,
  title={A transfer learning-based framework for enhanced classification of perceived mental stress using EEG spectrograms},
  author={Khan, Sheharyar and Noorani, Sadam Hussain and Frnda, Jaroslav and Rauf, Usman and Arsalan, Aamir and Saeed, Sanay Muhammad Umar},
  journal={IEEE Access},
  year={2025},
  publisher={IEEE}
}

@inproceedings{2_recommended_javed2024deep,
  title={A Deep Learning Approach for MDD Classification Using Frontal EEG Spectral Features with DenseNet-121},
  author={Javed, Muhammad Tariq and Ahmad, Waqar and Arsalan, Aamir and Khan, Sheharyar and Noorani, Sadam Hussain and Rauf, Usman},
  booktitle={2024 5th International Conference on Innovative Computing (ICIC)},
  pages={1--6},
  year={2024},
  organization={IEEE}
}

@article{3_recommended_khan2024machine,
  title={A machine learning based depression screening framework using temporal domain features of the electroencephalography signals},
  author={Khan, Sheharyar and Umar Saeed, Sanay Muhammad and Frnda, Jaroslav and Arsalan, Aamir and Amin, Rashid and Gantassi, Rahma and Noorani, Sadam Hussain},
  journal={Plos one},
  volume={19},
  number={3},
  pages={e0299127},
  year={2024},
  publisher={Public Library of Science San Francisco, CA USA}
}

@inproceedings{4_recommended_rehman2023eeg,
  title={EEG-based depression detection: a temporal domain feature-centric machine learning approach},
  author={Rehman, Maryam and Saeed, Sanay Muhammad Umar and Khan, Sheharyar and Noorani, Sadam Hussain and Rauf, Usman},
  booktitle={2023 International Conference on Frontiers of Information Technology (FIT)},
  pages={208--213},
  year={2023},
  organization={IEEE}
}

@article{benjamini2001control,
  title={The control of the false discovery rate in multiple testing under dependency},
  author={Benjamini, Yoav and Yekutieli, Daniel},
  journal={Annals of statistics},
  pages={1165--1188},
  year={2001},
  publisher={JSTOR}
}

@article{FPC_explainability_1,
  title={EEG differences between eyes-closed and eyes-open conditions at the resting stage for euthymic participants},
  author={Kan, DPX and Croarkin, PE and Phang, CK and Lee, PF},
  journal={Neurophysiology},
  volume={49},
  number={6},
  pages={432--440},
  year={2017},
  publisher={Springer}
}

@article{FPC_explainability_3,
  title={Eye closure causes widespread low-frequency power increase and focal gamma attenuation in the human electrocorticogram},
  author={Geller, Aaron S and Burke, John F and Sperling, Michael R and Sharan, Ashwini D and Litt, Brian and Baltuch, Gordon H and Lucas II, Timothy H and Kahana, Michael J},
  journal={Clinical Neurophysiology},
  volume={125},
  number={9},
  pages={1764--1773},
  year={2014},
  publisher={Elsevier}
}

@article{FPC_explainability_4,
  title={EEG differences between eyes-closed and eyes-open resting conditions},
  author={Barry, Robert J and Clarke, Adam R and Johnstone, Stuart J and Magee, Christopher A and Rushby, Jacqueline A},
  journal={Clinical neurophysiology},
  volume={118},
  number={12},
  pages={2765--2773},
  year={2007},
  publisher={Elsevier}
}

@ARTICLE{eegconformer,
  author={Song, Yonghao and Zheng, Qingqing and Liu, Bingchuan and Gao, Xiaorong},
  journal={IEEE Transactions on Neural Systems and Rehabilitation Engineering}, 
  title={EEG Conformer: Convolutional Transformer for EEG Decoding and Visualization}, 
  year={2023},
  volume={31},
  number={},
  pages={710-719},
  doi={10.1109/TNSRE.2022.3230250}}

@article{bci,
    author = {Cho, Hohyun and Ahn, Minkyu and Ahn, Sangtae and Kwon, Moonyoung and Jun, Sung Chan},
    title = {EEG datasets for motor imagery brain–computer interface},
    journal = {GigaScience},
    volume = {6},
    number = {7},
    pages = {gix034},
    year = {2017},
    month = {05},
   issn = {2047-217X},
    doi = {10.1093/gigascience/gix034}
}

@article{bandt2002permutation,
  title={Permutation entropy: a natural complexity measure for time series},
  author={Bandt, Christoph and Pompe, Bernd},
  journal={Physical review letters},
  volume={88},
  number={17},
  pages={174102},
  year={2002},
  publisher={APS}
}

@book{fossa2024occipital,
    author = {Abd-Elsayed, Alaa},
    title = {Basic Anesthesia Review},
    publisher = {Oxford University Press},
    year = {2024},
    month = {07},
    isbn = {9780197584569},
    doi = {10.1093/med/9780197584569.001.0001},
    url = {https://doi.org/10.1093/med/9780197584569.001.0001},
}

@article{posthoc_2022,
  title={A novel explainable machine learning approach for EEG-based brain-computer interface systems},
  author={Ieracitano, Cosimo and Mammone, Nadia and Hussain, Amir and Morabito, Francesco Carlo},
  journal={Neural Computing and Applications},
  volume={34},
  number={14},
  pages={11347--11360},
  year={2022},
  publisher={Springer}
}

@article{MNE_2013,
  title={MEG and EEG data analysis with MNE-Python},
  author={Gramfort, Alexandre and Luessi, Martin and Larson, Eric and Engemann, Denis A and Strohmeier, Daniel and Brodbeck, Christian and Goj, Roman and Jas, Mainak and Brooks, Teon and Parkkonen, Lauri and others},
  journal={Frontiers in Neuroinformatics},
  volume={7},
  pages={267},
  year={2013},
  publisher={Frontiers Media SA}
}

@article{pedregosa2011scikit,
  title={Scikit-learn: Machine learning in Python},
  author={Pedregosa, Fabian and Varoquaux, Ga{\"e}l and Gramfort, Alexandre and Michel, Vincent and Thirion, Bertrand and Grisel, Olivier and Blondel, Mathieu and Prettenhofer, Peter and Weiss, Ron and Dubourg, Vincent and others},
  journal={Journal of machine learning research},
  volume={12},
  number={Oct},
  pages={2825--2830},
  year={2011}
}

@article{bandpowerpaper,
  title={Dominant frequencies of resting human brain activity as measured by the electrocorticogram},
  author={Groppe, David M and Bickel, Stephan and Keller, Corey J and Jain, Sanjay K and Hwang, Sean T and Harden, Cynthia and Mehta, Ashesh D},
  journal={Neuroimage},
  volume={79},
  pages={223--233},
  year={2013},
  publisher={Elsevier}
}

@article {gomezram_eyes_op_close ,
	author = {G{\'o}mez-Ram{\'\i}rez, Jaime and Freedman, Shelagh and Mateos, Diego and P{\'e}rez-Vel{\'a}zquez, Jos{\'e} Luis and Valiante, Taufik},
	title = {Eyes closed or Eyes open? Exploring the alpha desynchronization hypothesis in resting state functional connectivity networks with intracranial EEG},
	elocation-id = {118174},
	year = {2017},
	doi = {10.1101/118174},
	publisher = {Cold Spring Harbor Laboratory},
	URL = {https://www.biorxiv.org/content/early/2017/03/18/118174},
	eprint = {https://www.biorxiv.org/content/early/2017/03/18/118174.full.pdf},
	journal = {bioRxiv}
}

@inproceedings{NEURIPS2020_c7bf0b7c,
 author = {Covert, Ian and Lundberg, Scott M and Lee, Su-In},
 booktitle = {Advances in Neural Information Processing Systems},
 editor = {H. Larochelle and M. Ranzato and R. Hadsell and M.F. Balcan and H. Lin},
 pages = {17212--17223},
 publisher = {Curran Associates, Inc.},
 title = {Understanding Global Feature Contributions With Additive Importance Measures},
 volume = {33},
 year = {2020}
}

@article{katsigiannis2017dreamer,
  title={DREAMER: A database for emotion recognition through EEG and ECG signals from wireless low-cost off-the-shelf devices},
  author={Katsigiannis, Stamos and Ramzan, Naeem},
  journal={IEEE journal of biomedical and health informatics},
  volume={22},
  number={1},
  pages={98--107},
  year={2017},
  publisher={IEEE}
}

@article{lim2018stew,
  title={STEW: Simultaneous task EEG workload data set},
  author={Lim, Wei Lun and Sourina, Olga and Wang, Lipo P},
  journal={IEEE Transactions on Neural Systems and Rehabilitation Engineering},
  volume={26},
  number={11},
  pages={2106--2114},
  year={2018},
  publisher={IEEE}
}

@article{williams2023crowdsourced,
  title={Crowdsourced EEG Experiments: A proof of concept for remote EEG acquisition using EmotivPRO Builder and EmotivLABS},
  author={Williams, Nikolas S and King, William and Mackellar, Geoffrey and Randeniya, Roshini and McCormick, Alicia and Badcock, Nicholas A},
  journal={Heliyon},
  volume={9},
  number={8},
  year={2023},
  publisher={Elsevier}
}

@INPROCEEDINGS{TUAB_dataset,
  author={López, S. and Suarez, G. and Jungreis, D. and Obeid, I. and Picone, J.},
  booktitle={2015 IEEE Signal Processing in Medicine and Biology Symposium (SPMB)}, 
  title={Automated identification of abnormal adult EEGs}, 
  year={2015},
  volume={},
  number={},
  pages={1-5},
  doi={10.1109/SPMB.2015.7405423}}

@INPROCEEDINGS{TUEV_dataset,
  author={Harati, A. and Golmohammadi, M. and Lopez, S. and Obeid, I. and Picone, J.},
  booktitle={2015 IEEE Signal Processing in Medicine and Biology Symposium (SPMB)}, 
  title={Improved EEG event classification using differential energy}, 
  year={2015},
  volume={},
  number={},
  pages={1-4},
  doi={10.1109/SPMB.2015.7405421}}

@article{fraiwan2012automated,
  title={Automated sleep stage identification system based on time--frequency analysis of a single EEG channel and random forest classifier},
  author={Fraiwan, Luay and Lweesy, Khaldon and Khasawneh, Natheer and Wenz, Heinrich and Dickhaus, Hartmut},
  journal={Computer methods and programs in biomedicine},
  volume={108},
  number={1},
  pages={10--19},
  year={2012},
  publisher={Elsevier}
}

@article{edla2018classification,
  title={Classification of EEG data for human mental state analysis using Random Forest Classifier},
  author={Edla, Damodar Reddy and Mangalorekar, Kunal and Dhavalikar, Gauri and Dodia, Shubham},
  journal={Procedia computer science},
  volume={132},
  pages={1523--1532},
  year={2018},
  publisher={Elsevier}
}

@article{christ2018time,
  title={Time series feature extraction on basis of scalable hypothesis tests (tsfresh--a python package)},
  author={Christ, Maximilian and Braun, Nils and Neuffer, Julius and Kempa-Liehr, Andreas W},
  journal={Neurocomputing},
  volume={307},
  pages={72--77},
  year={2018},
  publisher={Elsevier}
}

@article{yang2024biot,
  title={Biot: Biosignal transformer for cross-data learning in the wild},
  author={Yang, Chaoqi and Westover, M and Sun, Jimeng},
  journal={Advances in Neural Information Processing Systems},
  volume={36},
  year={2024}
}

@article{kostas2021bendr,
  title={BENDR: Using transformers and a contrastive self-supervised learning task to learn from massive amounts of EEG data},
  author={Kostas, Demetres and Aroca-Ouellette, Stephane and Rudzicz, Frank},
  journal={Frontiers in Human Neuroscience},
  volume={15},
  pages={653659},
  year={2021},
  publisher={Frontiers Media SA}
}

@article{weng2024self,
  title={Self-supervised learning for electroencephalogram: A systematic survey},
  author={Weng, Weining and Gu, Yang and Guo, Shuai and Ma, Yuan and Yang, Zhaohua and Liu, Yuchen and Chen, Yiqiang},
  journal={ACM Computing Surveys},
  volume={57},
  number={12},
  pages={1--38},
  year={2025},
  publisher={ACM New York, NY}
}

@article{schirrmeister2017deep,
  title={Deep learning with convolutional neural networks for EEG decoding and visualization},
  author={Schirrmeister, Robin Tibor and Springenberg, Jost Tobias and Fiederer, Lukas Dominique Josef and Glasstetter, Martin and Eggensperger, Katharina and Tangermann, Michael and Hutter, Frank and Burgard, Wolfram and Ball, Tonio},
  journal={Human brain mapping},
  volume={38},
  number={11},
  pages={5391--5420},
  year={2017},
  publisher={Wiley Online Library}
}

@article{mandhouj2021automated,
  title={An automated classification of EEG signals based on spectrogram and CNN for epilepsy diagnosis},
  author={Mandhouj, Badreddine and Cherni, Mohamed Ali and Sayadi, Mounir},
  journal={Analog integrated circuits and signal processing},
  volume={108},
  number={1},
  pages={101--110},
  year={2021},
  publisher={Springer}
}

@article{grandini2020metrics,
  title={Metrics for multi-class classification: an overview},
  author={Grandini, Margherita and Bagli, Enrico and Visani, Giorgio},
  journal={arXiv preprint arXiv:2008.05756},
  year={2020}
}

@article{li2022deep,
  title={A deep learning method approach for sleep stage classification with EEG spectrogram},
  author={Li, Chengfan and Qi, Yueyu and Ding, Xuehai and Zhao, Junjuan and Sang, Tian and Lee, Matthew},
  journal={International Journal of Environmental Research and Public Health},
  volume={19},
  number={10},
  pages={6322},
  year={2022},
  publisher={MDPI}
}

@article{chien2022maeeg,
  title={Maeeg: Masked auto-encoder for eeg representation learning},
  author={Chien, Hsiang-Yun Sherry and Goh, Hanlin and Sandino, Christopher M and Cheng, Joseph Y},
  journal={arXiv preprint arXiv:2211.02625},
  year={2022}
}

@article{oh2020deep,
  title={A deep learning approach for Parkinson’s disease diagnosis from EEG signals},
  author={Oh, Shu Lih and Hagiwara, Yuki and Raghavendra, U and Yuvaraj, Rajamanickam and Arunkumar, N and Murugappan, M and Acharya, U Rajendra},
  journal={Neural Computing and Applications},
  volume={32},
  pages={10927--10933},
  year={2020},
  publisher={Springer}
}

@inproceedings{mohammadi2024eeg2rep,
author = {Mohammadi Foumani, Navid and Mackellar, Geoffrey and Ghane, Soheila and Irtza, Saad and Nguyen, Nam and Salehi, Mahsa},
title = {EEG2Rep: Enhancing Self-supervised EEG Representation Through Informative Masked Inputs},
year = {2024},
isbn = {9798400704901},
publisher = {Association for Computing Machinery},
address = {New York, NY, USA},
url = {https://doi.org/10.1145/3637528.3671600},
doi = {10.1145/3637528.3671600},
booktitle = {Proceedings of the 30th ACM SIGKDD Conference on Knowledge Discovery and Data Mining},
pages = {5544–5555},
numpages = {12},
location = {Barcelona, Spain},
series = {KDD '24}
}

@article{chiarion2023connectivity,
  title={Connectivity Analysis in EEG Data: A Tutorial Review of the State of the Art and Emerging Trends},
  author={Chiarion, Giovanni and Sparacino, Laura and Antonacci, Yuri and Faes, Luca and Mesin, Luca},
  journal={Bioengineering},
  volume={10},
  number={3},
  pages={372},
  year={2023},
  publisher={MDPI}
}

@article{kuang2022phase,
  title={Phase lag index of resting-state eeg for identification of mild cognitive impairment patients with type 2 diabetes},
  author={Kuang, Yuxing and Wu, Ziyi and Xia, Rui and Li, Xingjie and Liu, Jun and Dai, Yalan and Wang, Dan and Chen, Shangjie},
  journal={Brain Sciences},
  volume={12},
  number={10},
  pages={1399},
  year={2022},
  publisher={MDPI}
}

@article{betrouni2019electroencephalography,
  title={Electroencephalography-based machine learning for cognitive profiling in Parkinson's disease: Preliminary results},
  author={Betrouni, Nacim and Delval, Arnaud and Chaton, Laurence and Defebvre, Luc and Duits, Annelien and Moonen, Anja and Leentjens, Albert FG and Dujardin, Kathy},
  journal={Movement Disorders},
  volume={34},
  number={2},
  pages={210--217},
  year={2019},
  publisher={Wiley Online Library}
}

@article{chaturvedi2017quantitative,
  title={Quantitative EEG (QEEG) measures differentiate Parkinson's disease (PD) patients from healthy controls (HC)},
  author={Chaturvedi, Menorca and Hatz, Florian and Gschwandtner, Ute and Bogaarts, Jan G and Meyer, Antonia and Fuhr, Peter and Roth, Volker},
  journal={Frontiers in aging neuroscience},
  volume={9},
  pages={3},
  year={2017},
  publisher={Frontiers Media SA}
}

@article{waninger2020neurophysiological,
  title={Neurophysiological biomarkers of Parkinson’s disease},
  author={Waninger, Shani and Berka, Chris and Stevanovic Karic, Marija and Korszen, Stephanie and Mozley, P David and Henchcliffe, Claire and Kang, Yeona and Hesterman, Jacob and Mangoubi, Tomer and Verma, Ajay},
  journal={Journal of Parkinson's disease},
  volume={10},
  number={2},
  pages={471--480},
  year={2020},
  publisher={IOS Press}
}

@article{lubba2019catch22,
  title={catch22: CAnonical Time-series CHaracteristics: Selected through highly comparative time-series analysis},
  author={Lubba, Carl H and Sethi, Sarab S and Knaute, Philip and Schultz, Simon R and Fulcher, Ben D and Jones, Nick S},
  journal={Data Mining and Knowledge Discovery},
  volume={33},
  number={6},
  pages={1821--1852},
  year={2019},
  publisher={Springer}
}

@article{abiri2019comprehensive,
  title={A comprehensive review of EEG-based brain--computer interface paradigms},
  author={Abiri, Reza and Borhani, Soheil and Sellers, Eric W and Jiang, Yang and Zhao, Xiaopeng},
  journal={Journal of neural engineering},
  volume={16},
  number={1},
  pages={011001},
  year={2019},
  publisher={IOP Publishing}
}

@article{casson2018electroencephalogram,
  title={Electroencephalogram},
  author={Casson, Alexander J and Abdulaal, Mohammed and Dulabh, Meera and Kohli, Siddharth and Krachunov, Sammy and Trimble, Eleanor},
  journal={Seamless healthcare monitoring: advancements in wearable, attachable, and invisible devices},
  pages={45--81},
  year={2018},
  publisher={Springer}
}

@incollection{sahota2024interpretable,
  title={Interpretable classification of early stage parkinson’s disease from eeg},
  author={Sahota, Amarpal and Roguski, Amber and Jones, Matthew W and Rolinski, Michal and Whone, Alan and Santos-Rodriguez, Raul and Abdallah, Zahraa S},
  booktitle={AI for Health Equity and Fairness: Leveraging AI to Address Social Determinants of Health},
  pages={219--231},
  year={2024},
  publisher={Springer}
}


\appendix

\section{KnowEEG Set Up} \label{appendix_knowEEG}
The TSFresh \cite{christ2018time} python package with version 0.20.3 was used to calculate the per electrode statistics with `Efficient' setting for feature calculation.  The MNE python package \cite{MNE_2013} version 1.7.1 was used to calculate the connectivity metrics, further details on connectivity metrics are provided in appendix \ref{appendix_connectivity}. The python package Scikit-learn \cite{pedregosa2011scikit} with version 1.6.1 was used to generate the trees of the Random Forest (with balanced sampling) . Decision trees with default hyperparameter settings were used for the Random Forest (criterion =`gini',  max depth=None, min samples split =2, min samples leaf = 1, max features = None).

\section{Functional Power Connectivity} \label{FPC}
In addition to correlation and conventional phase / coherence based connectivity measures we introduce an amplitude-based functional connectivity set of features which we term Functional Power Connectivity (FPC). For each electrode and frequency band (delta, theta, alpha, sigma, beta, gamma) we compute relative bandpower. We then define FPC features as the pairwise log-ratios of relative bandpower between electrodes.

Therefore for each band between each electrode pair i,j we compute :

\begin{equation}
\mathrm{FPC}_{i,j} = \log P^{\mathrm{rel}}_{i} - \log P^{\mathrm{rel}}_{j}
= \log\left( \frac{P^{\mathrm{rel}}_{i}}{P^{\mathrm{rel}}_{j}} \right)
\end{equation}

This means for a set of 14 electrodes there are in total 630 total FPC features comprising 6 x 14 band power features plus 91 x 6 bands of electrode-electrode features. These features capture regional band power as well as spatial differences in oscillatory power across regions which can be interpreted as power-based functional connectivity markers.

\section{Connectivity Metrics Overview} \label{appendix_connectivity}

Connectivity metrics for KnowEEG were selected using literature and are not exhaustive \cite{MNE_2013} \cite{kuang2022phase} . The selected metric is a hyperparameter of the pipeline. Therefore users can add metrics to their pipeline that they expect to be informative. 

Correlation is calculated in the time domain using the entire EEG signal. This results in a channel-channel matrix of features. As we calculate Pearson and Spearman correlation this leads to two channel-channel matrices of features. All other connectivity measures are calculated across the six power bands (delta, theta, alpha, sigma, beta, gamma). Therefore they result in six channel-channel matrices of features. For example for Phase Lag Index (PLI) this would lead to PLI on the delta band for each channel-channel pair, PLI on the theta band for each channel-channel pair etc...

Some metrics such as Phase Lag Index require segmentation of the EEG signal in order to determine how properties of the signal vary over time. Selected metrics for the pipeline are therefore split into those requiring and not requiring segmentation.

The MNE package was used for all connectivity metric calculations (except for correlation) and can therefore be referred to for exact definitions \cite{MNE_2013}. We use the same names below as in the MNE package.

Selected Metrics (no segmentation required) are: 
\begin{itemize}
    \item Correlation (Pearson and Spearman correlation, considered together as a single metric)
    \item Functional Power Connectivity 

\end{itemize}

Selected Metrics (with signal segmentation) are:
\begin{itemize}
    \item Coherence
    \item Imaginary Coherence
    \item Pairwise Phase Consistency 
    \item Phase Locking Value 
    \item Phase Lag Index
    \item Directed Phase Lag Index
    \item Weighted Phase Lag Index
\end{itemize}

In table \ref{tab:overview} below we show for the metrics with signal segmentation how the signal was segmented for each dataset. There is a trade-off between the length of each segment and number of segments in calculation of each metrics. More segments will lead to a more accurate calculation of the metric. However, a long enough segment is required in order to provide enough signal, particularly for lower frequency components of the signal. 

For DREAMER, STEW, and Crowdsourced due to the short nature of the signals, segmented metrics in the lower frequency bands are unlikely to contain useful information. However, for completeness we calculate them and allow KnowEEG to select from calculated metrics.

% \begin{table}[ht]
% \renewcommand{\arraystretch}{1.4} 
%     \centering
%     \resizebox{\columnwidth}{!}{%
%         \begin{tabular}{lcccc}
%             \hline
%             \textbf{Dataset} & \textbf{Signal Duration} & \textbf{Freq} & \textbf{Segment Duration} & \textbf{Num Segments} \\
%             \hline
%             DREAMER       & 2s   & 128Hz & 0.25  & 8  \\
%             STEW          & 2s   & 128Hz & 0.25  & 8  \\
%             Crowdsourced  & 2s   & 128Hz & 0.25  & 8  \\
%             TUEV          & 5s   & 256Hz & 0.50  & 10 \\
%             TUAB          & 10s  & 256Hz & 0.625 & 16 \\
%             \hline
%         \end{tabular}
%     }
%     \caption{Overview of segmentation for signals for each dataset. Required in order to calculate specific connectivity metrics (Pairwise Phase Consistency, Phase Locking Value, Phase Lag Index, Directed Phase Lag Index, Weighted Phase Lag Index).}
%     \label{tab:overview}
% \renewcommand{\arraystretch}{1} 
% \end{table}

\begin{table}[ht]
\renewcommand{\arraystretch}{1.4} 
    \centering
    \resizebox{\columnwidth}{!}{%
        \begin{tabular}{lcccc}
            \hline
            \textbf{Dataset} & \makecell{\textbf{Signal} \\ \textbf{Duration}} & \textbf{Freq} & \makecell{\textbf{Segment} \\ \textbf{Duration}} & \textbf{Num Segments} \\
            \hline
            DREAMER       & 2s   & 128Hz & 0.25  & 8  \\
            STEW          & 2s   & 128Hz & 0.25  & 8  \\
            Crowdsourced  & 2s   & 128Hz & 0.25  & 8  \\
            TUEV          & 5s   & 256Hz & 0.50  & 10 \\
            TUAB          & 10s  & 256Hz & 0.625 & 16 \\
            \hline
        \end{tabular}
    }
    \caption{Overview of segmentation for signals for each dataset. Required in order to calculate specific connectivity metrics (Coherence, Imaginary Coherence, Pairwise Phase Consistency, Phase Locking Value, Phase Lag Index, Directed Phase Lag Index, Weighted Phase Lag Index).}
    \label{tab:overview}
\renewcommand{\arraystretch}{1} 
\end{table}

\section{Datasets} \label{datasets}
Datasets and preprocessing follow the same protocol as in \cite{mohammadi2024eeg2rep}.

\subsection{Emotiv datasets}
All Emotiv datasets were bandpass filtered and windowed into segments of length 256. This data is recorded at a sampling rate of 128Hz. Therefore, each segment is of 2 seconds in length.

\subsubsection*{\textbf{DREAMER}} 

 The DREAMER \cite{katsigiannis2017dreamer} dataset is a multimodal database containing both electroencephalogram (EEG) and electrocardiogram (ECG) signals. These signals are recorded during affect elicitation with audi-visual stimuli \cite{katsigiannis2017dreamer}. Data is recorded from 23 participants along with self-assessment of affective state after each stimuli. Self assessment is in terms of valence, arousal and dominance. For classification we use the arousal labels as per \cite{mohammadi2024eeg2rep}. We utilize the toolkit Torcheeg for preprocessing which consists of low-pass and high-filters. We do not use the ECG data and use only the EEG data. The DREAMER dataset can be accessed here\footnote{https://zenodo.org/records/546113}.

\subsubsection*{\textbf{Crowdsource} }
The Crowdsourced \cite{gomezram_eyes_op_close} dataset was recorded with participants at rest in the eyes closed or eyes open state. Each recording is 2 minutes in total. There are 60 total participants with only 13 recording data for both states. Data is recorded using 14-channel EPOC devices and is initially recorded at 2048Hz and then downsampled to 128Hz. This data can be accessed via the Open Science Framework \footnote{https://osf.io/9bvgh} 

\subsubsection*{\textbf{Simultaneous Task EEG Workload (STEW)}}
The STEW dataset \cite{lim2018stew} is an open access EEG dataset for multitasking and mental workload analysis. There are 48 total subjects and data is recorded using a 14-channel Emotiv EPOC headset.  Data is recorded for subjects at baseline (rest) and under workload in the SIMKAP mutitasking setting. EEG recordings are of length 2.5 minutes and recorded at 128Hz. Participants recorded their perceived mental workload on a scale of 1 to 9. STEW also has binary labels recording workload of above 4 as high and below 4 as low. These binary labels are used for our classification task. STEW is accessiible via the IEEE DataPort \footnote{6https://ieee-dataport.org/open-access/stew-simultaneous-task-eeg-workloaddataset}.

\subsection{Temple University Hospital (TUH) Datasets}
The TUH EEG Events dataset (TUEV) \cite{TUEV_dataset} and the TUH Abnormal EEG corpus (TUAB) \cite{TUAB_dataset} are accessible on request from the Neural Engineering Data Consortium (NEDC) here  \footnote{https://isip.piconepress.com/projects/nedc/html/tuh\_eeg/}. The TUH datasets were processed in accordance with standard 16 EEG montages \cite{mohammadi2024eeg2rep} and the 10-20 international system with: 
"FP1-F7", "F7-T7", "T7-P7", "P7-O1", "FP2-F8", "F8-T8", "T8-P8", "P8-O2", "FP1-F3", "F3-C3",
"C3-P3", "P3-O1", "FP2-F4", "F4-C4", "C4-P4", and "P4-O2".

\subsubsection*{\textbf{TUH Events Corpus (TUEV)}}
The TUH EEG Events Corpus (TUEV) \cite{TUEV_dataset} \cite{mohammadi2024eeg2rep} contains EEG data with samples segmented into six categories. The categories are spike and sharp wave, eye movement, artifact, background generalized periodic epileptiform discharges and periodic lateralized epileptiform discharges \cite{TUEV_dataset}.

\subsubsection*{\textbf{TUH Abnormal EEG Corpus (TUAB)}}
The TUH Abnormal EEG Corpus (TUAB) \cite{TUAB_dataset} \cite{mohammadi2024eeg2rep} is a large-scale corpus of EEGs recordings designed to support research on automated EEG interpretation. This corpus is drawn from a diverse group of subjects with EEG labeled as either normal or abnormal \cite{TUAB_dataset}.

\section{Computational Complexity Analysis}
\label{appendix:complexity}

In this section, we provide a theoretical comparison of the computational demands of KnowEEG versus the deep learning baselines, accounting for feature extraction, tuning, and model training. We also provide by example using the Crowdsource dataset peak memory requirements reached during training.

\subsection{Terminology and Variable Definitions}
Table \ref{tab:complexity_vars} defines the parameters used to quantify the computational complexity.

\begin{table}[h]
\centering
\caption{Definitions for Complexity Analysis}
\label{tab:complexity_vars}
\begin{tabular}{lll}
\hline
\textbf{Symbol} & \textbf{Definition} & \textbf{Context} \\ \hline
$N$ & Number of samples & Total dataset size \\
$K$ & Number of channels & EEG electrodes (14--16) \\
$L$ & Sequence length & Data points per sample \\
$M$ & Total features & $A$ (per-electrode) + $B$ (connectivity) \\
$T$ & Number of trees & Random Forest hyperparameter \\
$C$ & Candidate metrics & Connectivity metrics for local selection \\
$H$ & Tuning iterations & Hyperparameter search space size \\
$E$ & Training epochs & Iterative updates (Deep Learning) \\
$d$ & Model dimension & Hidden layer size (Deep Learning) \\ \hline
\end{tabular}
\end{table}

\subsection{Theoretical Comparison}
As shown in Table \ref{tab:complexity_comp}, KnowEEG provides efficiency advantages through its non-iterative training and local tuning strategies.

\begin{table}[h]
\centering
\caption{Complexity and Memory Comparison: KnowEEG vs. Deep Learning Baselines}
\label{tab:complexity_comp}
\resizebox{\textwidth}{!}{
\begin{tabular}{llll}
\hline
\textbf{Metric} & \textbf{KnowEEG} & \textbf{EEG Conformer} & \textbf{EEG2Rep} \\ \hline
Feature Extraction & $O(N \cdot K \cdot L \log L)$ & $O(N \cdot K \cdot L)$ & $O(N \cdot K \cdot L)$ \\
Tuning (Search) & $O(C \cdot N \log N)$ & $O(H \cdot E \cdot N \cdot L^2 \cdot d)$ & $O(H \cdot E \cdot N \cdot L^2 \cdot d)$ \\
Training Time & $O(T \cdot \sqrt{M} \cdot N \log N)$ & $O(E \cdot N \cdot (L \cdot d + L^2 \cdot d))$ & $O(E \cdot N \cdot L^2 \cdot d)$ \\
Peak Memory & $O(N \cdot M)$ & $O(L^2 + \text{weights})$ & $O(L^2 + \text{weights})$ \\
Hardware & CPU & GPU (High VRAM) & GPU (High VRAM) \\ \hline
\end{tabular}
}
\end{table}

\subsection{Analysis Summary}
A significant practical advantage of KnowEEG is that once deterministic feature extraction is completed, the training time and peak memory requirements are entirely independent of the sequence length ($L$). In contrast, Transformer-based architectures such as EEG2Rep and EEG Conformer incur a quadratic cost ($O(L^2)$) for both computation and memory during the self-attention phase. This necessitates substantial GPU VRAM to store activation maps across $E$ training epochs. We show by practical example in table \ref{tab:empirical_resource_comp} on the Crowdsource dataset, how the memory requirements for training varied across the KnowEEG, EEG Conformer and EEG2Rep models.

\begin{table}[h]
\centering
\caption{KnowEEG vs. Deep Learning Baselines memory requirements on Crowdsourced Dataset}
\label{tab:empirical_resource_comp}
\resizebox{\textwidth}{!}{
\begin{tabular}{llll}
\hline
\textbf{Performance Metric} & \textbf{KnowEEG} & \textbf{EEG Conformer} & \textbf{EEG2Rep} \\ \hline
Peak  Memory  & 3 GB (CPU RAM) & 8 GB (GPU VRAM) & 10 GB (GPU VRAM) \\
Memory Type & CPU RAM & Dedicated GPU VRAM & Dedicated GPU VRAM \\
Training Overhead & Exactly 1 Training Pass & E Epochs & E Epochs + Pre-training
\\ \hline
\end{tabular}
}
\end{table}

KnowEEG also uses \textit{local parameter selection} for connectivity metrics, avoiding a global grid search across the $M$-dimensional feature space. While the Random Forest requires tuning the number of trees ($T$), this is a one-pass operation per tree, unlike the iterative back-propagation required for models like EEG2Rep. Finally, self supervised representation-based models such as EEG2Rep require an intensive pre-training phase that KnowEEG bypasses entirely while remaining competitive in performance.

\end{document}